\documentclass[journal]{IEEEtran}

\usepackage{cite}

\ifCLASSINFOpdf
  \usepackage[pdftex]{graphicx}
\else
\fi
\usepackage{amsmath}
\usepackage{latexsym} 
\usepackage{amssymb}

\usepackage{algorithmic}

\usepackage{array}

  \usepackage[caption=false,font=normalsize,labelfont=sf,textfont=sf,labelformat=simple]{subfig}

\newcommand{\etal}{\textit{et al}.}
\newcommand{\ie}{\textit{i}.\textit{e}.}
\newcommand{\eg}{\textit{e}.\textit{g}.}
\newcommand{\etc}{\textit{etc}}

\begin{document}
%
\title{Perceptual Adversarial Networks for Image-to-Image Transformation}
%
%
%

\author{Chaoyue~Wang,
             Chang~Xu,
             Chaohui~Wang,
             Dacheng~Tao 
\thanks{This work was supported
by the Australian Research Council Projects: FL-170100117, DE-180101438, DP-180103424, and LP-150100671. This
work was partially supported by SAP SE and CNRS INS2IJCJC-INVISANA.}
\thanks{Chaoyue Wang is with the Centre for Artificial Intelligence, School of Software, the Faculty of Engineering and Information Technologies, University of Technology Sydney and the UBTECH Sydney Artificial Intelligence Centre and the School of Information Technologies, the Faculty of Engineering and Information Technologies, the University of Sydney, Australia (e-mail: chaoyue.wang@student.uts.edu.au)}
\thanks{Chang Xu and Dacheng Tao are with the UBTECH Sydney Artificial Intelligence Centre and the School of Information Technologies, the Faculty of Engineering and Information Technologies, the University of Sydney, 6 Cleveland St, Darlington, NSW 2008, Australia (email: {c.xu, dacheng.tao}@sydney.edu.au).}
\thanks{Chaohui Wang is with the Universit\'{e} Paris-Est, LIGM Lab, UMR 8049 CNRS$\cdot$ENPC$\cdot$ESIEE$\cdot$UPEM, Marne-la-Vall\'{e}e, France (e-mail: chaohui.wang@u-pem.fr)}
\thanks{\textcopyright 20XX IEEE. Personal use of this material is permitted.
Permission from IEEE must be obtained for all other uses, in any current or future media, including reprinting/republishing this material for advertising or promotional purposes, creating new collective works, for resale or redistribution to servers or lists, or reuse of any copyrighted component of this work in other works.}}

\maketitle

\begin{abstract}
In this paper, we propose Perceptual Adversarial Networks (PAN) for image-to-image transformations. Different from existing application driven algorithms, PAN provides a generic framework of learning to map from input images to desired images (Fig.~\ref{Demo}), such as a rainy image to its de-rained counterpart, object edges to photos, semantic labels to a scenes image,~\etc. The proposed PAN consists of two feed-forward convolutional neural networks (CNNs): the image transformation network $T$ and the discriminative network $D$. Besides the generative adversarial loss widely used in GANs, we propose the perceptual adversarial loss, which undergoes an adversarial training process between the image transformation network $T$ and the hidden layers of the discriminative network $D$. The hidden layers and the output of the discriminative network $D$ are upgraded to constantly and automatically discover the discrepancy between the transformed image and the corresponding ground-truth, while the image transformation network $T$ is trained to minimize the discrepancy explored by the discriminative network $D$.  Through integrating the generative adversarial loss and the perceptual adversarial loss, $D$ and $T$ can be trained alternately to solve image-to-image transformation tasks. Experiments evaluated on several image-to-image transformation tasks (\eg, image de-raining, image inpainting, \etc) demonstrate the effectiveness of the proposed PAN and its advantages over many existing works.
\end{abstract}


%
\IEEEpeerreviewmaketitle

\section{Introduction}
\label{sec:introduction}

\IEEEPARstart{I}{mage-to-image} transformations aim to transform an input image into the desired output image, and they exist in a number of applications about image processing, computer graphics, and computer vision. For example, generating high-quality images from corresponding degraded (e.g. simplified, corrupted or low-resolution) images, and transforming a color input image into its semantic or geometric representations. More examples include, but not limited to, image de-noising~\cite{elad2006image}, image in-painting~\cite{bertalmio2000image}, image super-resolution~\cite{nasrollahi2014super}, image colorization~\cite{luan2007natural}, image segmentation~\cite{khan2014survey}, \etc. 

In recent years, convolutional neural networks (CNNs) are trained in a supervised manner for various image-to-image transformation tasks~\cite{fu2017clearing, pathak2016context, dong2016image, zhang2016colorful}. They encode input image into hidden representation, which is then decoded to the output image. By penalizing the discrepancy between the output image and ground-truth image, optimal CNNs can be trained to discover the mapping from the input image to the transformed image of interest. These CNNs are developed with distinct motivations and differ in the loss function design.

One of the most straightforward approaches is to pixel-wisely evaluate output images~\cite{dong2016image,zhang2016colorful, cheng2015deep,shelhamer2016fully,du2018quantum}, \eg, least squares or least absolute losses to calculate the distance between the output and ground-truth images in the pixel space. Though pixel-wise evaluation can generate reasonable images, there are some unignorable defects associated with the outputs, such as image blur and image artifacts. 

Besides pixel-wise losses, the generative adversarial losses were largely utilized in training image-to-image transformation models. GANs (and cGANs)~\cite{goodfellow2014generative,miyato2018cgans} perform an adversarial training process alternating between identifying and faking, and generative adversarial losses are formulated to evaluate the discrepancy between the generated distribution and the real-world distribution. Experimental results show that generative adversarial losses are beneficial for generating more realistic images. Therefore, there are many GANs (or cGANs) based works to solve image-to-image transformation tasks, resulting in sharper and more realistic transformed images~\cite{pathak2016context,isola2016image}. Meanwhile, some GANs variants~\cite{Zhu_2017_ICCV,Yi_2017_ICCV,pmlr-v70-kim17a,chen2018attention} investigated cross-domain image translations and performed image translations in absence of paired examples. Although these unpaired works achieved reasonable results in some image-to-image translation tasks, they are inappropriate for some image-to-image problems. For example, in image in-painting tasks, it is difficult to define the domain and formulate the distribution of corrupted images. In addition, paired information within training data are beneficial for learning image transformations, but they cannot be utilized by unpaired translation methods.

\begin{figure*}[!t]
\centering
\includegraphics[width=7in]{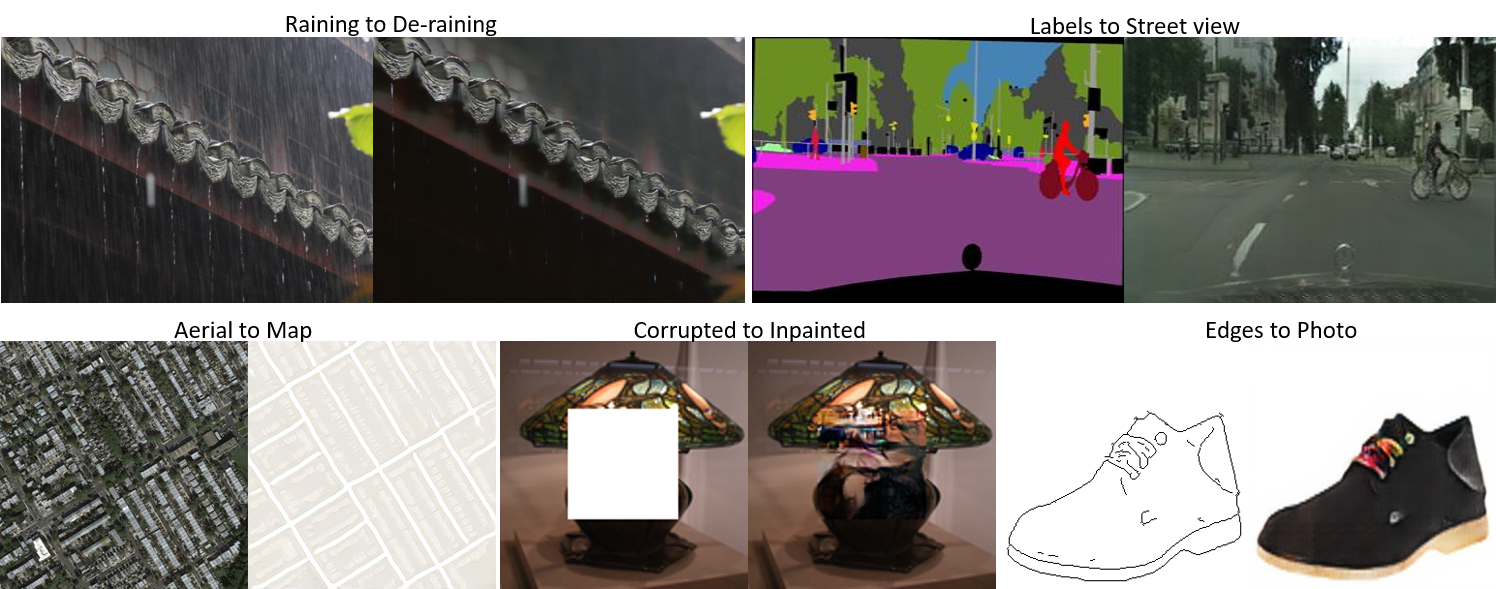}
\caption{Image-to-image transformation tasks. Many tasks in image processing, computer graphics, and computer vision can be regarded as image-to-image transformation tasks, where a model is designed to transform an input image into the required output image. We proposed Perceptual Adversarial Networks (PAN) to solve the image-to-image transformation between paired images. For each pair of the images we demonstrated, the left one is the input image, and the right one is the transformed result of the proposed PAN.}
\label{Demo}
\end{figure*} 

Moreover, perceptual losses emerged as a novel measurement for evaluating the discrepancy between high-level perceptual features of the output and ground-truth images~\cite{johnson2016perceptual, dosovitskiy2016generating, bruna2015super}. Hidden layers of a well-trained image classification network (\eg, VGG-16~\cite{simonyan2014very}) are usually employed to extract high-level features (\eg, content or texture) of both output images and ground-truth images. It is then expected to encourage the output image to have the similar high-level feature with that of the ground-truth image. Recently, perceptual losses were introduced in aforementioned GANs-based image-to-image transformation frameworks for suppressing artifacts~\cite{zhang2017image} and improving perceptual quality~\cite{ledig2016photo,zhang2018unreasonable} of the output images. Though integrating perceptual losses into GANs has produced impressive image-to-image transformation results, existing works are used to depend on external well-trained image classification network (\eg VGG-Net) out of GANs, but ignored the fact that GANs, especially the discriminative network, also has the capability and demand of perceiving the content of images and the difference between images. Moreover, since these external networks are trained on specific classification datasets (\eg, ImageNet), they mainly focus on features that contribute to the classification and may perform inferior in some image transformation tasks (\eg, transfer aerial images to maps). Meanwhile, since specific hidden layers of pre-trained networks are employed, it is difficult to explore the difference between generated images and ground-truth images from more points of view.

In this paper, we proposed the perceptual adversarial networks (PAN) for image-to-image transformation tasks. Inspired by GANs, PAN is composed of an image transformation network $T$ and a discriminative network $D$. Both generative adversarial loss and perceptual adversarial loss are employed. Firstly, similar with GANs, the generative adversarial loss is utilized to measure the distribution of generated images, \ie, penalizing generated images to lie in the desired target domain, which usually contributes to producing more visually realistic images. Meanwhile, to comprehensively evaluate transformed images, we devised the perceptual adversarial loss to form dynamic measurements based on the hidden layers of the discriminative network $D$. Specifically, given hidden layers of the network $D$, the network $T$ is trained to generate the output image that has the same high-level features with that of the corresponding ground-truth. If the difference between images measured on existing hidden layers of the discriminator is smaller, these hidden layers will be updated to discover the discrepancy between images from a new point of view. Different from the pixel-wise loss and conventional perceptual loss, our perceptual adversarial loss undergoes an adversarial training process, and aims to discover and decrease the discrepancy under constantly explored dynamic measurements. 

In summary, our paper makes the following contributions: 

\begin{itemize}
\item We proposed the perceptual adversarial loss, which utilizes the hidden layers of the discriminative network to evaluate the discrepancy between the output and ground-truth images through an adversarial training process. 

\item Through combining the perceptual adversarial loss and the generative adversarial loss, we presented the PAN for solving image-to-image transformation tasks.

\item We evaluated the performance of the PAN on several image-to-image transformation tasks (Fig.~\ref{Demo}). Experimental results show that the proposed PAN has a great capability of accomplishing image-to-image transformations. 
\end{itemize}

The rest of the paper is organized as follows: after a brief summary of previous related works in section~\ref{sec:background}, we illustrate the proposed PAN together with its training losses in section~\ref{sec:methods}. Then we exhibit the experimental validation of the whole method in section~\ref{sec:experiments}. Finally, we conclude this paper with some future directions in section~\ref{sec:conclusion}.

\section{Background}
\label{sec:background}

In this section, we first introduce some representative image-to-image transformation methods based on feed-forward CNNs, and then summarize related works on GANs and perceptual losses. 

\subsection{Image-to-image transformation with feed-forward CNNs}

Recent years have witnessed a variety of feed-forward CNNs developed for image-to-image transformation tasks. These feed-forward CNNs can be easily trained using the back-propagation algorithm~\cite{rumelhart1988learning}, and the transformed images are generated by forwardly passing the input image through the well-trained CNNs in the test stage. 

Individual pixel-wise loss or pixel-wise loss accompanied with other losses are employed in a number of image-to-image transformations. Image super-resolution tasks estimate a high-resolution image from its low-resolution counterpart~\cite{dong2016image, johnson2016perceptual, ledig2016photo}. Image de-raining (or de-snowing) methods attempt to remove the rain (or snow) strikes in the pictures brought by the uncontrollable weather conditions~\cite{eigen2013restoring,fu2017clearing,zhang2017image}. Given a damaged image, image inpainting aims to recover the missing part of the input image~\cite{pathak2016context,ruzic2015context,qin2014novel}. Image semantic segmentation methods produce dense scene labels based on a single input image~\cite{farabet2013learning, noh2015learning, eigen2015predicting}. Given an input object image, some feed-forward CNNs were trained to synthesize the image of the same object from a different viewpoint~\cite{yang2015weakly,wang2017tag}. More image-to-image transformation tasks based on feed-forward CNNs, include, but not limited to, image colorization~\cite{cheng2015deep}, depth estimations~\cite{eigen2014depth,eigen2015predicting}, \etc.

\subsection{GANs-based works}
Generative adversarial networks (GANs)~\cite{goodfellow2014generative} provide an important approach for learning a generative model which generates samples from the real-world data distribution. GANs consist of a generative network and a discriminative network. Through playing a minimax game between these two networks, GANs are trained to generate more and more ‘realistic’ samples. Since the great performance on learning real-world distributions, there have emerged a large number of GANs-based works. Some of these GANs-based works are committed to training a better generative model, such as InfoGAN~\cite{chen2016infogan}, Energy-based GAN~\cite{zhao2016energy}, WGAN(-GP)~\cite{arjovsky2017wasserstein,gulrajani2017improved},  Progressive GAN~\cite{karras2017progressive}, E-GAN~\cite{wang2018evolutionary} and SN-GAN~\cite{miyato2018spectral}. There are also some works integrating the GANs into their models to improve the performance of classical tasks. For example,  the PGAN~\cite{Li_2017_CVPR} is proposed for small object detection. Specifically,~\cite{Li_2017_CVPR} devised a novel perceptual discriminator network, which contains an adversarial branch and a perception branch. The adversarial branch utilizes the adversarial loss to distinguish representations of real and synthesized objectives; the perception branch (or loss) employs a classification loss $L_{cls}$ and a bounding-box regression loss $L_{loc}$ to encourage the synthesized `super-resolved` objective’s representation to retain the same perception information as the input small objective’s representation.

In addition, these kind of works include, but not limited to, the PGN~\cite{lotter2015unsupervised} for video prediction, the SRGAN~\cite{ledig2016photo} for super-resolution, the ID-CGAN for image de-raining~\cite{zhang2017image}, the iGAN~\cite{zhu2016generative} for interactive application, the IAN~\cite{brock2016neural} for photo modification, and the Context-Encoder for image in-painting~\cite{pathak2016context}. Most recently, Isola~\etal~\cite{isola2016image} proposed the pix2pix-cGANs to perform several image-to-image transformation tasks (also known as image-to-image translations in their work), such as translating semantic labels into the street scene, object edges into pictures, aerial photos into maps, \etc.

Moreover, some GANs variants~\cite{Zhu_2017_ICCV,Yi_2017_ICCV,pmlr-v70-kim17a} investigated cross-domain image translations through exploring the cyclic mapping (or primal-dual) relation between different image domains. Specifically, a primal GAN aims to explore the mapping relations from source images to target images, while a dual (or inverse) GAN performs the invert task. These two GANs form a closed loop and allow images from either domain to be translated and then reconstructed. Through combining the GAN loss and cycle consistency loss (or recovery loss), these works can be used for performing image translation tasks in absence of paired examples. 
However, if paired training data are available in some applications, \cite{Zhu_2017_ICCV,Yi_2017_ICCV,pmlr-v70-kim17a} neglect paired information between data often have inferior performance to that of paired methods~\cite{isola2016image}.  Thus, at this stage, it is still important to study paired training, especially for performance-driven situations and applications, such as high-resolution image synthesis~\cite{wang2017high}, photo-realistic image synthesis~\cite{ledig2016photo}, real-world image inpainting~\cite{chen2017photographic}, \etc.

\subsection{Perceptual loss}
Recently, some theoretical analysis and experimental results suggested that the high-level features extracted from a well-trained image classification network have the capability to capture the perceptual information from real-world images~\cite{gatys2015neural,johnson2016perceptual}. Specifically, representations extracted from hidden layers of well-trained image classification network are beneficial to interpret the semantics of input images, and image style distribution can be captured by the Gram matrix of hidden representations. Hence, high-level features extracted from hidden layers of a well-trained classifier are often introduced in image generation models. Dosovitskiy and Brox~\cite{dosovitskiy2016generating} took Euclidean distances between high-level features of images as the deep perceptual similarity metrics to improve the performance of image generation. Johnson \etal~\cite{johnson2016perceptual}, Bruna \etal~\cite{bruna2015super} and Ledig \etal~\cite{ledig2016photo} used features extracted from a well-trained VGG network to improve the performance of single image super-resolution task. In addition, there are works applying high-level features in image style-transfer~\cite{gatys2015neural, johnson2016perceptual}, image de-raining~\cite{zhang2017image} and image view synthesis~\cite{park2017transformation} tasks.

\begin{figure*}[!t]
\centering
\includegraphics[width=6in]{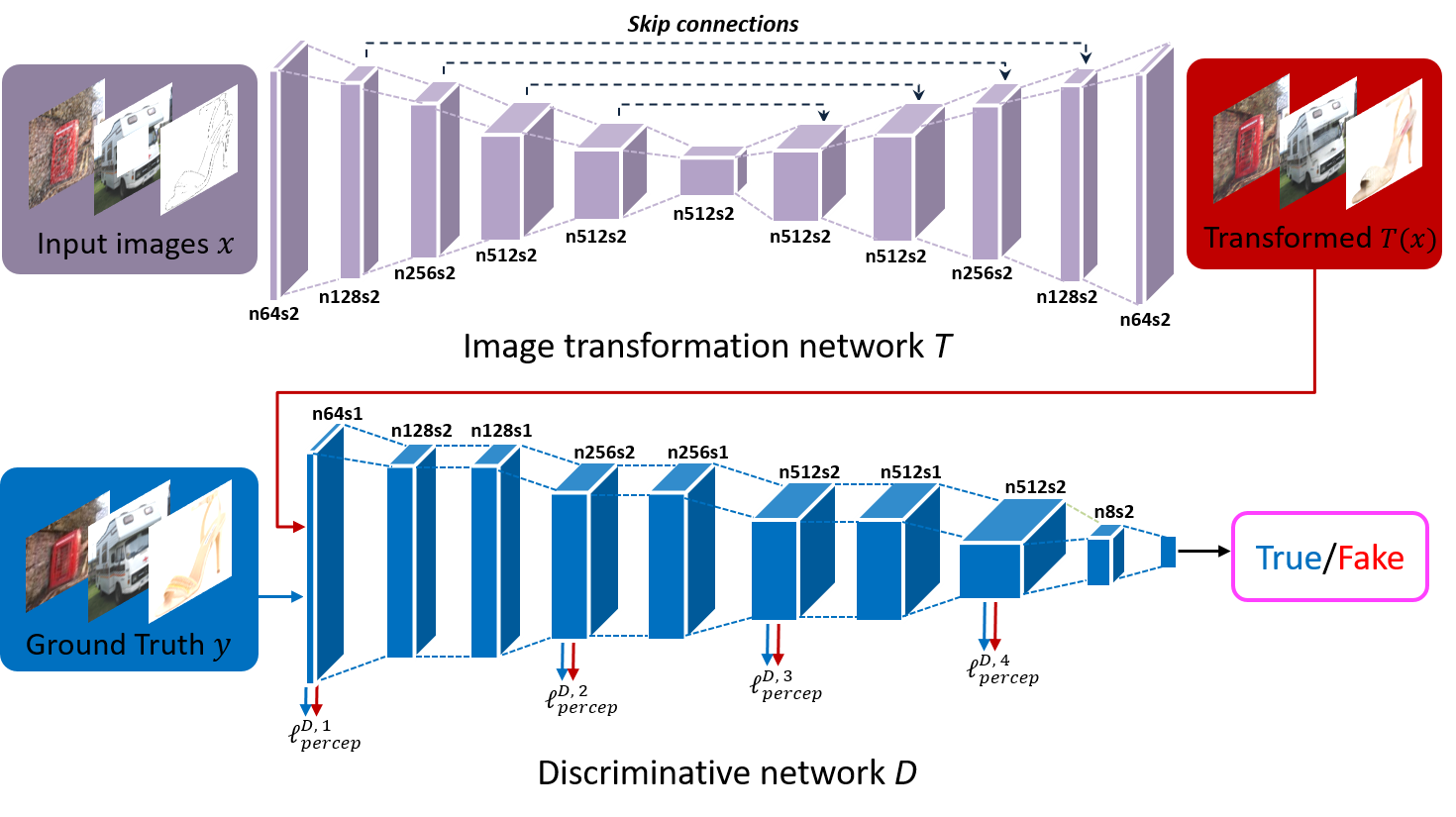} \vskip -0.2 in
\caption{PAN framework. PAN consists of an image transformation network $T$ and a discriminative network $D$. The image transformation network $T$ is trained to synthesize the transformed images given the input images. It is composed of a stack of Convolution-BatchNorm-LeakyReLU encoding layers and Deconvolution-BatchNorm-ReLU decoding layers, and the skip-connections are used between mirrored layers. The discriminative network $D$ is also a CNN that consists of Convolution-BatchNorm-LeakyReLU layers. Hidden layers of the network $D$ are utilized to evaluate the perceptual adversarial loss, and the output of the network $D$ is used to distinguish transformed images from real-world images.}
\label{framework}
\end{figure*}

\section{Methods}
\label{sec:methods}
In this section, we introduce the proposed Perceptual Adversarial Networks (PAN) for image-to-image transformation tasks. Firstly, we explain the generative and perceptual adversarial losses, respectively. Then, we give the whole framework of the proposed PAN. Finally, we illustrate the details of the training procedure and network architectures. 

\subsection{Generative adversarial loss}
We begin with the generative adversarial loss in vanilla GANs. A generative network $G$ is trained to map samples from noise distribution $p_z$ to real-world data distribution $p_\text{data}$ through playing a minimax game with a discriminative network $D$. In the training procedure, the discriminative network $D$ aims to distinguish the real samples $y \sim p_\text{data}$ from the generated samples $G(z)$. In contrary, the generative network $G$ tries to confuse the discriminative network $D$ by generating increasingly realistic samples. This minimax game can be formulated as:
\begin{equation}
\begin{aligned} 
\min_G \max_D \mathbb{E}_{y \sim p_\text{data}}[\log D(y)] +\mathbb{E}_{z \sim p_z} [\log (1-D(G(z)))]
\end{aligned} 
\end{equation} 

Nowadays, GANs-based models have shown the strong capability of learning generative models, especially for image generation~\cite{arjovsky2017wasserstein, lotter2015unsupervised, chen2016infogan}. We, therefore, adopt the GANs learning strategy to solve image-to-image transformation tasks as well. As shown in Fig.~\ref{framework}, the image transformation network $T$ is used to generate transformed image $T(x)$ given the input image $x \in \mathcal{X}$. Meanwhile, each input image $x$ has a corresponding ground-truth image $y$. 
We suppose that all target image $y \in \mathcal{Y}$ obey the distribution $p_\text{real}$, and the transformed image $T(x)$ is encouraged to have the same distribution with that of targets image $y$, \ie, $T(x) \sim p_\text{real}$. To achieve the generative adversarial learning strategy, a discriminative network $D$ is additionally introduced, and the generative adversarial loss can be written as:
\begin{equation}
\begin{aligned} 
\min_T \max_D \mathcal{V}_{D,T} = \mathbb{E}_{y \in \mathcal{Y}}[\log D(y)]  + \mathbb{E}_{x \in \mathcal{X}} [\log (1-D(T(x)))]
\end{aligned} \label{eq:gal}
\end{equation} 
The generative adversarial loss acts as a statistical measurement to penalize the discrepancy between the distributions of transformed images and the ground-truth images. 

\subsection{Perceptual adversarial loss}
Different from vanilla GANs that randomly generate samples from the data distribution $p_\text{data}$, our goal is to infer the transformed image according to the input images. Therefore, it is a further step of GANs to explore the mapping from the input image to its ground truth.

As mentioned in Sections~\ref{sec:introduction} and~\ref{sec:background}, pixel-wise losses and perceptual losses are widely used in existing works for generating images towards the ground truth. The pixel-wise losses penalize the discrepancy occurred in the pixel space, but often produce blurry results~\cite{pathak2016context, zhang2016colorful}. The perceptual losses explore the discrepancy between high-dimensional representations of images extracted from a well-trained classifier, \eg, the VGG net trained on the ImageNet dataset~\cite{simonyan2014very}. Although hidden layers of well-trained classifier have been experimentally validated to map the image from pixel space to high-level feature spaces, how to extract the effective features for image-to-image transformation tasks from hidden layers has not been thoroughly discussed. 

\begin{figure*}[!t]
\centering
\includegraphics[bb=2000bp 110bp 1400bp 1100bp,scale=0.157]{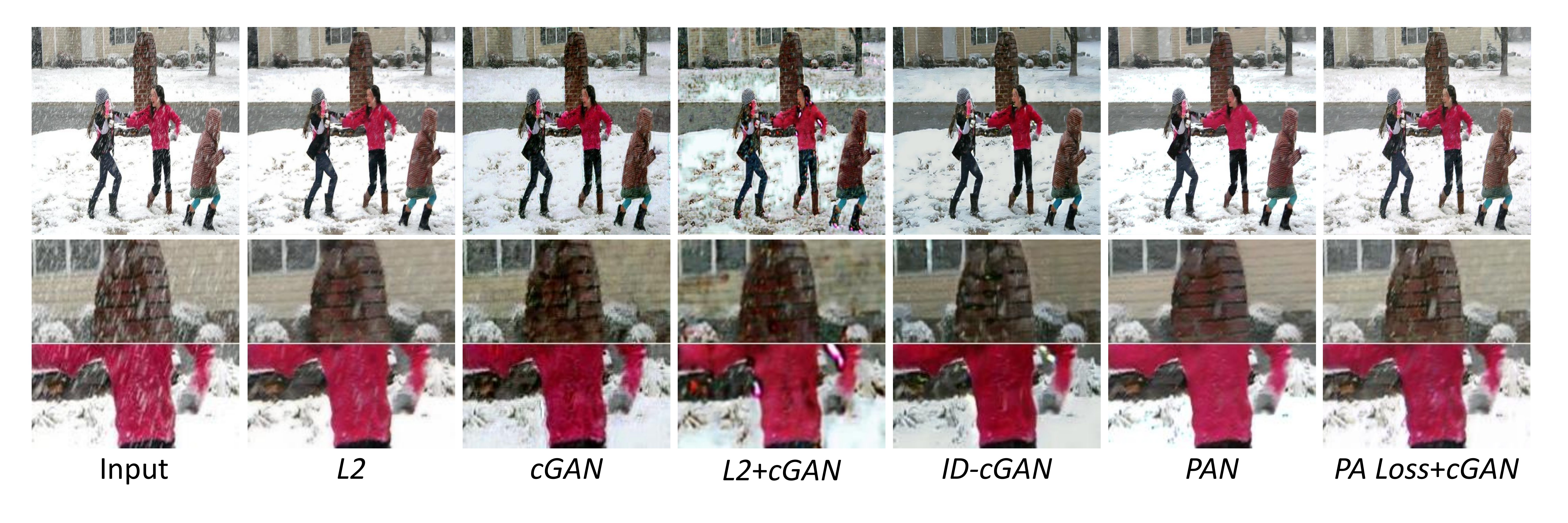}
\caption{Comparison of snow-streak removal using different losses functions. Given the same input image (leftmost), each column shows results trained under different losses. The loss function of ID-CGAN~\cite{zhang2017image} combined the pixel-wise loss (least squares loss), cGANs loss and perceptual loss, \ie, L2+cGAN+perceptual. For better visual comparison, zoomed versions of the specific regions-of-interest are demonstrated below the test images.}
\label{derain_real}
\end{figure*}

Here, we employ hidden layers of the discriminative network $D$ to evaluate the perceptual adversarial loss between transformed images and ground-truth images. In our experiments, given the training sample $\{(x_i,y_i) \in (\mathcal{X}\times \mathcal{Y})\}_{i=1}^{N}$, the least absolute loss is employed to calculate the discrepancy of the high-dimensional representations on the hidden layers of the network $D$, \eg, 
\begin{equation}
\ell_{percep}^{D, j} = \frac{1}{N}\sum_{i=1}^{N}||d_j(y_i) - d_j(T(x_i)) ||
\end{equation} 
where $d_j(\cdot)$ is the image representation on the $j$\textsuperscript{th} hidden layer of the discriminative network $D$, and $\ell_{percep}^{D, j}$ calculates the discrepancy measured by the $j$\textsuperscript{th} hidden layer of $D$.

Similar to what has been done with the Energy-Based GAN~\cite{zhao2016energy}, we use two different losses, one ($\mathcal{L}_T$) to train the image transformation network $T$, and the other ($\mathcal{L}_{D}$) to train hidden layers of the discriminative network $D$. Therefore, the image transformation network $T$ and hidden layers of the discriminative network $D$ play a non-zero-sum game and form the perceptual adversarial loss. Formally, the perceptual adversarial loss $\mathcal{L}_T$ for the image transformation network $T$ can be written as:
\begin{equation}
\mathcal{L}_T = \sum_{j=1}^F \lambda_j \ell_{percep}^{D, j} \label{eq:palt}
\end{equation}
and, given a positive margin $m$, the loss $\mathcal{L}_{D}$ for hidden layers of the discriminative network $D$ is defined as:
\begin{equation}
\begin{aligned} 
\mathcal{L}_{D} & = [m - \mathcal{L}_T]^+ = \Big[m - \sum_{j=1}^F \lambda_j\ell_{percep}^{D, j} \Big]^+ \label{eq:pald}
\end{aligned} 
\end{equation} 
where $[\cdot]^+=\max(0,\cdot)$, $\{\lambda_j\}_{j=1}^F$ are hyper-parameters balancing the influence of $F$ different hidden layers. 

By minimizing the perceptual adversarial loss function $\mathcal{L}_T$ with respect to parameters of $T$, we encourage the network $T$ to generate image $T(x)$ that has similar high-level features with its ground-truth $y$ on the hidden layers. If the weighted sum of discrepancy between transformed images and ground-truth images on different hidden layers is less than the positive margin $m$, the loss function $\mathcal{L}_{D}$ will upgrade the discriminative network $D$ for some new latent feature spaces, which preserve the discrepancy between the transformed images and their ground-truth. Therefore, based on the perceptual adversarial loss, the discrepancy between the transformed and ground-truth images can be constantly explored and exploited.

Compared to our perceptual adversarial loss which measures the difference between the transformed image and ground-truth image in hidden layers of the discriminator, the conditional GAN loss indicates whether the transformed image forms the appropriate image pair with the input image, and can also explore supervised information of paired images during the training process. However, they utilize different methods to minimize high-level feature differences explored by the discriminator. The perceptual adversarial loss directly penalizes the high-level representations of transformed images and ground-truth images to be as same as possible. In contrast, the conditional GAN loss aims to model the mapping relation from the input $x$ to its output $y_\text{real}$ and encourages the generated image pairs $(x,T(x))$ obeying the same conditional distribution $P_\text{real}(y|x)$. Compared to conditional GAN loss that indirectly guides the generated images $T(x)$ sharing the same features with corresponding ground-truth $y_\text{real}$, our perceptual adversarial loss directly measures and minimizes differences between generated images and ground-truth images from different perspectives. 

\subsection{The perceptual adversarial networks}

Based on the aforementioned generative adversarial loss (Eq.~\ref{eq:gal}) and perceptual adversarial loss (Eq.~\ref{eq:palt} and Eq.~\ref{eq:pald}), we develop the PAN framework, which consists of an image transformation network $T$ and a discriminative network $D$. These two networks are trained alternately to perform an adversarial learning process, the loss functions of image transformation network $\mathcal{J}_T$ and discriminative network $\mathcal{J}_D$ are formally defined as:
\begin{equation}
\begin{aligned} 
\mathcal{J}_T & = \theta\mathcal{V}_{D,T} + \mathcal{L}_T\\
\mathcal{J}_D & = -\theta\mathcal{V}_{D,T} + \mathcal{L}_{D}\\
& = -\theta\mathcal{V}_{D,T} + [m - \mathcal{L}_{T}]^+
\end{aligned}
\end{equation}
where $\theta$ is the hyper-parameter balance the influence of generative adversarial loss and perceptual adversarial loss. When $\mathcal{L}_{T}<m$, minimizing $\mathcal{J}_D$ with respect to the parameters of $D$ is consistent with maximizing $\mathcal{J}_T$. Otherwise, when $\mathcal{L}_{T} \geq m$, the second term of $\mathcal{J}_D$ will have zero gradients, because of the positive margin $m$.
In general, the discriminative network $D$ aims to distinguish transformed image $T(x)$ from ground-truth image $y$ from both the statical (the first term of $\mathcal{J}_D$) and dynamic perceptual (the second term of $\mathcal{J}_D$) aspects. On the other hand, the image transformation network $T$ is trained to generate increasingly better images by reducing the discrepancy between the output and ground-truth images.

\begin{figure*}[!t]
\centering
\includegraphics[width=7.1in]{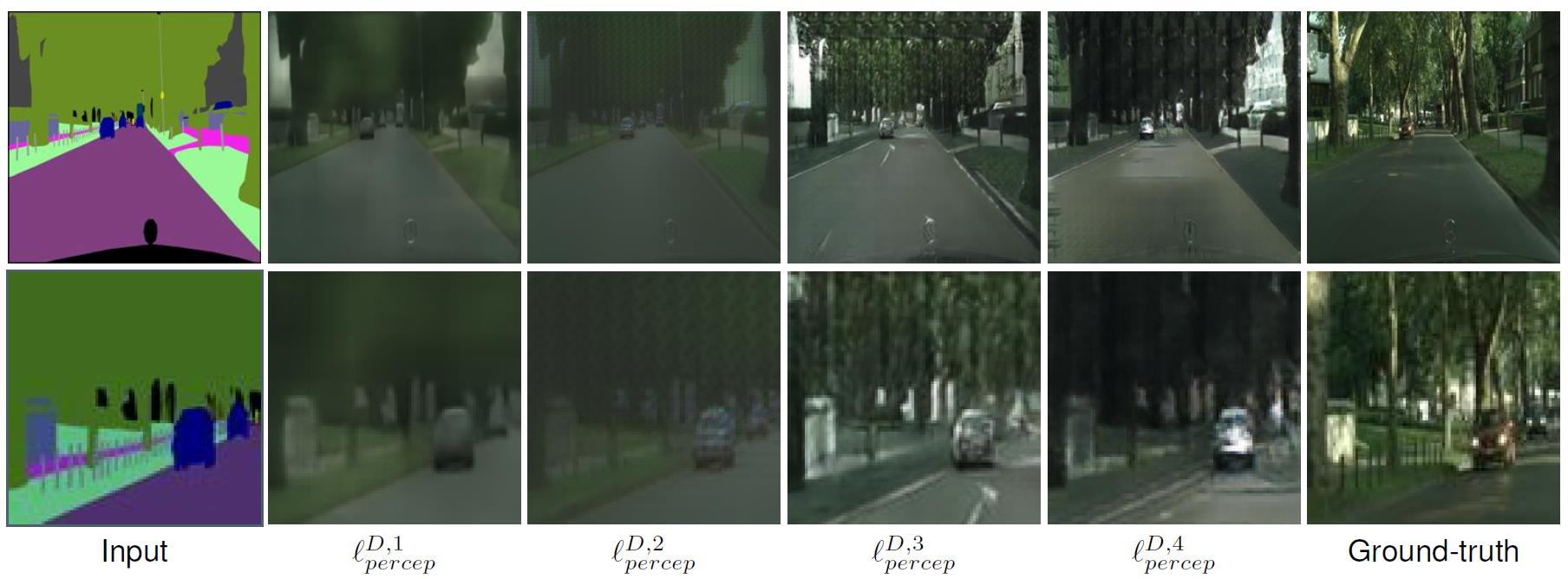}
\caption{Transforming the semantic labels to cityscapes images use the perceptual adversarial loss. Within the perceptual adversarial loss, a different hidden layer is utilized for each experiment. For better visual comparison, zoomed versions of the specific regions-of-interest are demonstrated below the test images. For higher layers, the transformed images look sharper, but less color information is preserved..}
\label{fig:analysis} 
\end{figure*}

\subsection{Network architectures}

Fig.~\ref{framework} illustrates the framework of the proposed PAN, which is composed of two CNNs, \ie, the image transformation network $T$ and the discriminative network $D$. 

\subsubsection{Image transformation network $T$}

The image transformation network $T$ is designed to generate the transformed image given the input image. Following the network architectures in~\cite{radford2015unsupervised,isola2016image}, the network $T$ firstly encodes the input image into high-dimensional representation using a stack of Convolution-BatchNorm-LeakyReLU layers, and then, the output image can be decoded by the following Deconvolution-BatchNorm-ReLU layers\footnote{The deconvolution layer utilized in our framework is the transposed convolution layer used in~\cite{zeiler2011adaptive,dumoulin2016guide}.}. Note that the output layer of the network $T$ does not use batchnorm and replaces the ReLU with Tanh activation. Moreover, the skip-connections are used to connect mirrored layers in the encoder and decoder stacks. More details of the transformation network $T$ are listed in Table~\ref{table:netG}. The same architecture of the network $T$ is used for all experiments in this paper, except there is an additional explanation\footnote{In the analysis of the loss functions and the image inpainting task, different architectures of the network $T$ were used.}.

\begin{figure*}[!t]
\centering
\includegraphics[bb=100bp 40bp 1590bp 460bp,scale=0.32]{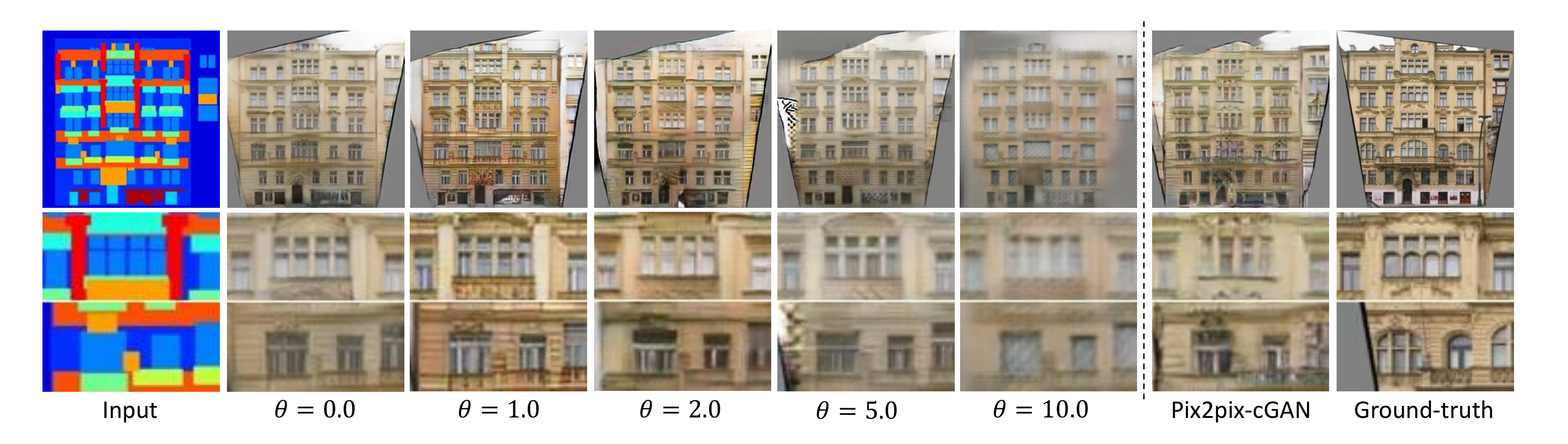}
\caption{Comparison of transforming the semantic labels to facades images by controlling the hyper-parameter $\theta$. Given the same input image (leftmost), each column shows results trained under different $\theta$. For better visual comparison, zoomed versions of the specific regions-of-interest are demonstrated below the test images.}
\label{facades} 
\end{figure*}

\begin{table}[!t]
\renewcommand{\arraystretch}{1.5}
\caption{The architecture of the discriminative network.}
\label{table:netD}
\centering
\begin{tabular}{l}
\hline
\ \ \ \ \ \ \ \ \ \ \  \ \ \ \ \ \ \ \ \ \ \ \ {\bf Discriminative network} $D$\\
\hline
{\bf Input}: Image  \\
\hline
[layer 1]    \ \ \ Conv. (3, 3, 64), stride=1; \emph{LReLU}; \\
\hline
\ \ \ \ \ \ \ \ \ \ \ \ \ \ \big(Perceptual adversarial loss: $\ell_{percep}^{D, 1}$\big)\\
\hline
[layer 2]    \ \ \  Conv. (3, 3, 128), stride=2; Batchnorm; \emph{LReLU};     \ \ \ \ \  \\
\hline
[layer 3]    \ \ \  Conv. (3, 3, 128), stride=1; Batchnorm; \emph{LReLU};     \ \ \ \ \  \\
\hline
[layer 4]    \ \ \ Conv. (3, 3, 256), stride=2; Batchnorm; \emph{LReLU};\\
\hline
\ \ \ \ \ \ \ \ \ \ \ \ \ \ \big(Perceptual adversarial loss: $\ell_{percep}^{D, 2}$\big)\\
\hline
[layer 5]    \ \ \  Conv. (3, 3, 256), stride=1; Batchnorm; \emph{LReLU};     \ \ \ \ \  \\
\hline
[layer 6]    \ \ \ Conv. (3, 3, 512), stride=2; Batchnorm; \emph{LReLU};\\
\hline
 \ \ \ \ \ \ \ \ \ \ \ \ \ \ \big(Perceptual adversarial loss: $\ell_{percep}^{D, 3}$\big)\\
\hline
[layer 7]    \ \ \  Conv. (3, 3, 512), stride=1; Batchnorm; \emph{LReLU};     \ \ \ \ \\
\hline
[layer 8]    \ \ \  Conv. (3, 3, 512), stride=2; Batchnorm; \emph{LReLU};\\
\hline
\ \ \ \ \ \ \ \ \ \ \ \ \ \  \big(Perceptual adversarial loss: $\ell_{percep}^{D, 4}$\big)\\
\hline
[layer 9]     \ \ \  Conv. (3, 3, 8), stride=2; \emph{LReLU};     \ \ \ \ \  \\
\hline
[layer 10]    \ \  Fully connected (1); \emph{Sigmoid};    \ \ \ \ \  \\
\hline
{\bf Output}: Real or Fake (Probability)\\
\hline
\end{tabular}
\end{table}

\begin{table}[!t]
\renewcommand{\arraystretch}{1.5}
\caption{The architecture of the image transformation network.}
\label{table:netG}
\centering
\begin{tabular}{l}
\hline
\ \ \ \ \ \ \ \ \ \ \ \ \ \ \ \ \ \ \ \ \ \ \ \ {\bf Image transformation network} $T$\\
\hline
{\bf Input}: Image \\
\hline
[layer 1]    \ \ \ \ \ Conv. (3, 3, \ 64), stride=2; \emph{LReLU};     \ \ \ \ \ \\
\hline
[layer 2]    \ \ \ \ \  Conv. (3, 3, 128), stride=2; Batchnorm; \\
\hline
[layer 3]    \ \ \  \ \  \emph{LReLU}; Conv. (3, 3, 256), stride=2; Batchnorm; \\
\hline
[layer 4]    \ \ \ \ \   \emph{LReLU}; Conv. (3, 3, 512), stride=2; Batchnorm;\\
\hline
[layer 5]    \ \ \ \ \   \emph{LReLU}; Conv. (3, 3, 512), stride=2; Batchnorm;\\
\hline
[layer 6]    \ \ \ \ \  \emph{LReLU}; Conv. (3, 3, 512), stride=2; Batchnorm; \emph{LReLU};     \ \ \ \\
\hline
[layer 7]    \ \ \ \ \  DeConv. (4, 4, 512), stride=2; Batchnorm; \     \ \ \  \\
\hline
\ \ \ \ \ \ \ \ \ \ \ \ \ \ \ \ Concatenate Layer(Layer 7, Layer 5); \emph{ReLU}; \ \ \ \  \\
\hline
[layer 8]    \ \ \ \ \  DeConv. (4, 4, 256), stride=2; Batchnorm; \\
\hline
\ \ \ \ \ \ \ \ \ \ \ \ \ \ \ \ Concatenate Layer(Layer 8, Layer 4); \emph{ReLU}; \ \ \ \  \\
\hline
[layer 9]    \ \ \ \ \  DeConv. (4, 4, 128), stride=2; Batchnorm;  \ \ \  \\
\hline
\ \ \ \ \ \ \ \ \ \ \ \ \ \ \ \ Concatenate Layer(Layer 9, Layer 3); \emph{ReLU}; \ \ \ \  \\
\hline
[layer 10]    \ \ \ \  DeConv. (4, 4, \ 64), stride=2; Batchnorm; \\
\hline
\ \ \ \ \ \ \ \ \ \ \ \ \ \ \ \ Concatenate Layer(Layer 10, Layer 2); \emph{ReLU}; \ \ \ \  \\
\hline
[layer 11]    \ \ \ \  DeConv. (4, 4, \ 64), stride=2; Batchnorm; \emph{ReLU}; \\
\hline
[layer 12]  \ \ \ \  DeConv. (4, 4, 3), stride=2; \emph{Tanh};     \ \ \ \ \ \\
\hline
{\bf Output}: Transformed image\\
\hline
\end{tabular}
\end{table}

\subsubsection{Discriminative network $D$}
In the proposed PAN framework, the discriminative network $D$ is introduced to compute the discrepancy between the transformed images and the ground-truth images. Specifically, given an input image, the discriminative network $D$ extracts high-level features using a series of Convolution-BatchNorm-LeakyReLU layers. The 1\textsuperscript{st}, 4\textsuperscript{th}, 6\textsuperscript{th}, and 8\textsuperscript{th} layers are utilized to measure the perceptual adversarial loss for every pair of transformed image and its corresponding ground-truth in the training data. Finally, the last convolution layer is flattened and then fed into a single sigmoid output. The output of the discriminative network $D$ estimates the probability that the input image comes from the real-world dataset rather than from the image transformation network $T$. 
The same discriminative network $D$ is applied for all tasks demonstrated in this paper, and details of the network $D$ are shown in Table~\ref{table:netD}.

\section{Experiments}
\label{sec:experiments}

In this section, we evaluate the performance of the proposed PAN on several image-to-image transformation tasks, which are popular in fields of image processing (\eg, image de-raining), computer vision (\eg, semantic segmentation) and computer graphics (\eg, image generation).

\subsection{Experimental setting up}
\label{sec:setup}
For fair comparisons, we adopted the same settings with existing works, and reported experimental results using several evaluation metrics. These tasks and data settings include:

\begin{itemize}
\item \emph{Single image de-raining}, on the dataset provided by ID-CGAN~\cite{zhang2017image}. 
\item \emph{Image Inpainting}, on a subset of ILSVRC'12 (same as context-encoder~\cite{pathak2016context}).
\item \emph{Semantic labels$\leftrightarrow$images}, on the Cityscapes dataset~\cite{cordts2016cityscapes} (same as pix2pix~\cite{isola2016image}).
\item \emph{Edges$\to$images}, on the dataset created by pix2pix~\cite{isola2016image}. The original data is from~\cite{zhu2016generative} and~\cite{yu2014fine}, and the HED edge detector~\cite{xie2015holistically} was used to extract edges.
\item \emph{Aerial$\to$map}, on the dataset from pix2pix~\cite{isola2016image}.
\end{itemize}

\begin{figure*}[!t]
\centering
\includegraphics[width=7.1in]{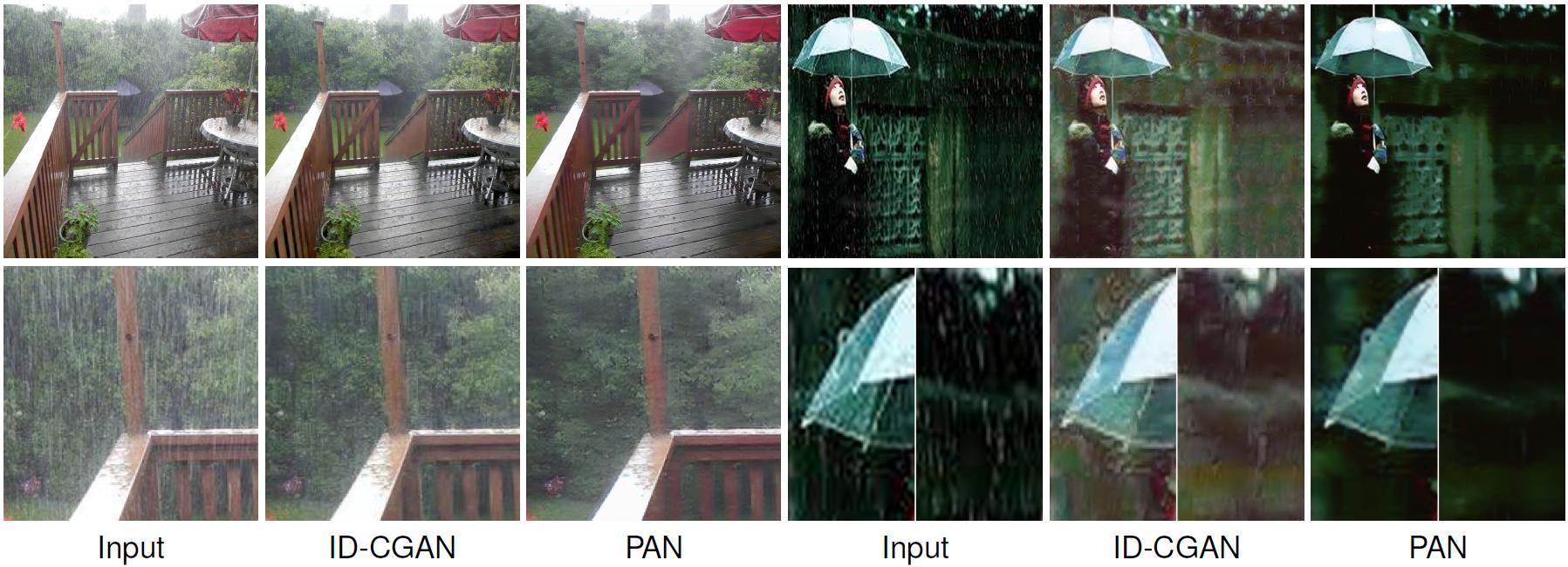}
\caption{Comparison of rain-streak removal using the ID-CGAN with the proposed PAN on real-world rainy images. For better visual comparison, zoomed versions of the specific regions-of-interest are demonstrated below the test images.}
\label{derain_idcgan} 
\end{figure*}

Furthermore, all experiments were trained on Nvidia Titan-X GPUs using Theano~\cite{bergstra2010theano}. Given the generative and perceptual adversarial losses, we alternately updated the image transformation network $T$ and the discriminative network $D$. Specifically, Adam solver~\cite{kingma2014adam} with a learning rate of 0.0002 and a first momentum of 0.5 was used in network training. After one update of the discriminative network $D$, the image transformation $T$ will be updated three times. Hyper-parameters $\theta=1$, $\lambda_1=5$, $\lambda_2=1.5$, $\lambda_3=1.5$, $\lambda_4=1$, and batch size of 4 were used for all tasks. Since the dataset sizes for different tasks are changed largely, the training epochs of different tasks were set accordingly. Overall, the number of training iterations was around 100k.

\subsection{Evaluation metrics}

To illustrate the performance of image-to-image transformation tasks, we conducted qualitative and quantitative experiments to evaluate the performance of the transformed images. For the qualitative experiments, we directly presented the input and transformed images. Meanwhile, we used quantitative measures to evaluate the performance over the test sets, such as Peak Signal to Noise Ratio (PSNR), Structural Similarity Index (SSIM)~\cite{wang2004image}, Universal Quality Index (UQI)~\cite{995823} and Visual Information Fidelity (VIF)~\cite{sheikh2006image}. f training iterations were around 100k.

\begin{figure}[!t]
\centering
\includegraphics[width=3.5in]{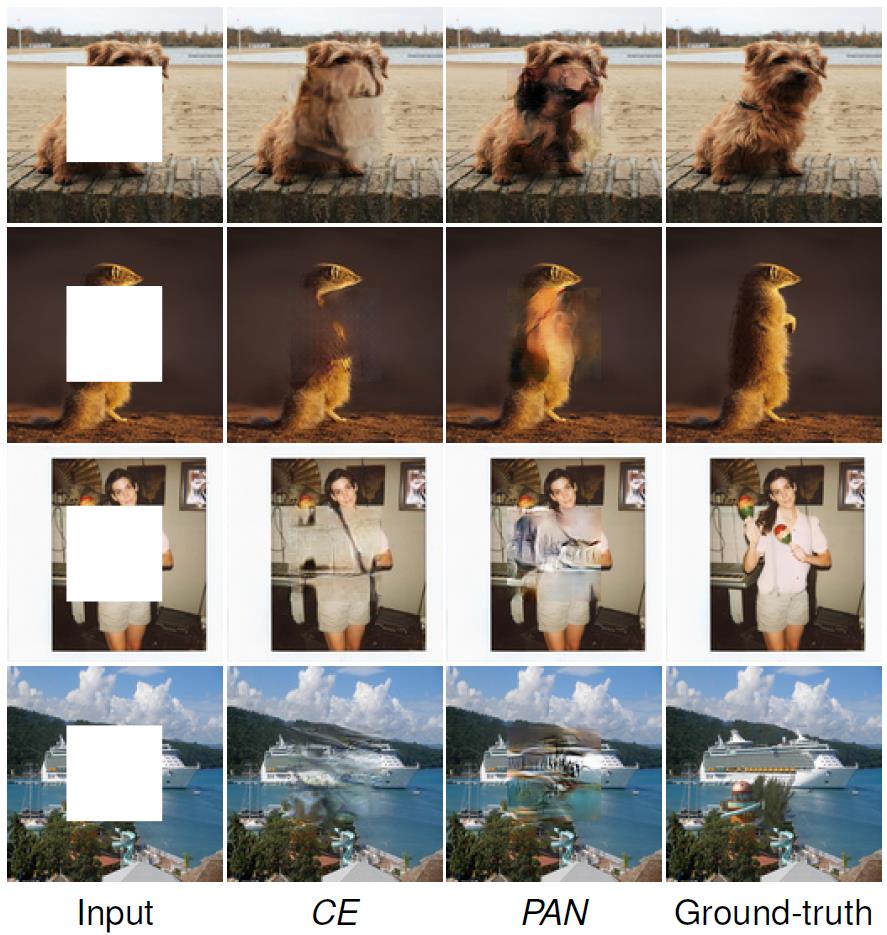}
\caption{Comparison of image in-painting results using the Context-Encoder(CE) with the proposed PAN. Given the central region missed input image (leftmost), the in-painted images and the ground-truth are listed on its rightside.}
\label{fig_sim}  
\end{figure}

\begin{figure*}[!t]
\centering
\includegraphics[width=7.1in]{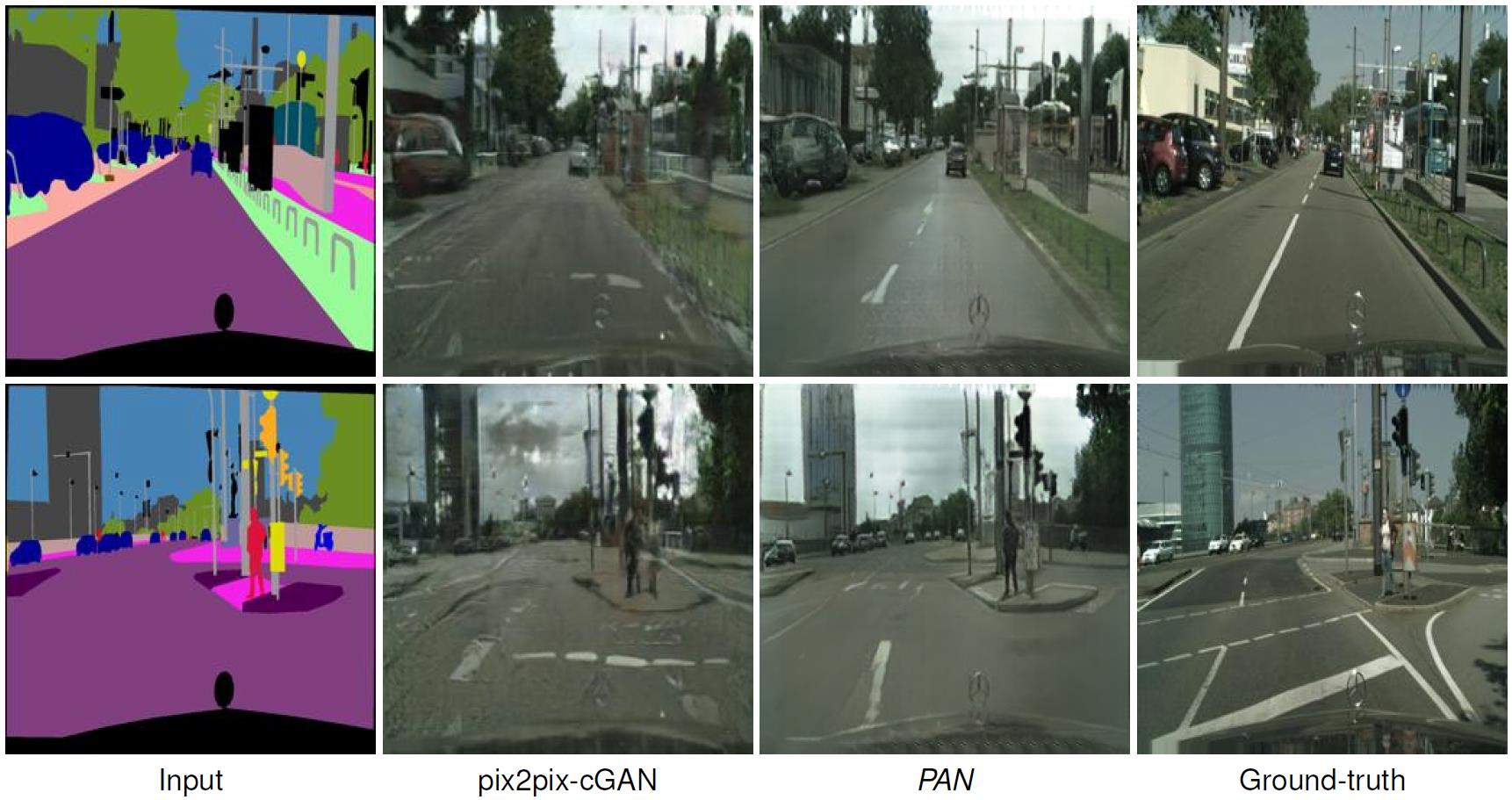}
\caption{Comparison of transforming the semantic labels to cityscapes images using the pix2pix-cGAN with the proposed PAN. Given the semantic labels (leftmost), the transformed cityscapes images and the ground-truth are listed on the rightside.}
\label{cityscapes2} 
\end{figure*}

\subsection{Analysis of the loss functions}

As discussed in Sections~\ref{sec:introduction} and~\ref{sec:background}, the design of loss function will largely influence the performance of image-to-image transformation. Firstly, the pixel-wise loss (using least squares loss) is widely used in various image-to-image transformation works~\cite{dong2014learning,kingma2013auto}. Then, the joint loss integrating pixel-wise loss and conditional generative adversarial loss is proposed to synthesize more realistic transformed images~\cite{pathak2016context,isola2016image}. Most recently, through introducing the perceptual loss, \ie, penalizing the discrepancy between high-level features that extracted by a well-trained classifier, the performance of some image-to-image transformation tasks are further enhanced~\cite{ledig2016photo,johnson2016perceptual,zhang2017image}. Different from these existing methods, the proposed PAN loss integrates the generative adversarial loss and the perceptual adversarial loss to train image-to-image transformation networks.
Here, we compare the performance of the proposed perceptual adversarial loss with those of existing losses. For a fair comparison, we adopted the same image transformation network and data settings from ID-CGAN~\cite{zhang2017image}, and used the combination of different losses to perform the image de-raining (de-snowing) task. The quantitative results over the synthetic test set were shown in Table~\ref{table:deraining}, while the qualitative results on the real-world images were shown in Fig.~\ref{derain_real}. 
From both quantitative and qualitative comparisons, we find that only using the pixel-wise loss (least squares loss) achieved the worst result, and there are many snow-streaks in the transformed images (Fig.~\ref{derain_real}). Through introducing the cGANs loss, the de-snowing performance was indeed improved, but artifacts can be observed (Fig.~\ref{derain_real}) and the PSNR performance dropped (Table~\ref{table:deraining}). Combining the pixel-wise,  cGAN and perceptual loss (VGG-16~\cite{simonyan2014very}) together, \ie, using the loss function of ID-CGAN~\cite{zhang2017image}, the quality of transformed images has been further improved on both observations and quantitative measurements. However, from Fig.~\ref{derain_real}, we observe that the transformed images have some color distortion compared to the input images. 
The proposed PAN loss (i.e., combining the perceptual adversarial loss and original GAN loss) not only removed most streaks without color distortion, but also achieved much better performance on quantitative measurements. Moreover, we evaluated the performance of combining conditional GAN loss and the perceptual adversarial loss. Comparing with using the cGAN loss independently, introducing the perceptual adversarial loss largely improves the model performance. Yet, comparing with the PAN loss, replacing the original GAN loss with its conditional version does not make a further improvement in both quantitative and qualitative comparisons.

Though variables of the discriminator network are optimized in iterations, the capability of hidden layers is constrained by network architecture. Therefore, in the proposed PAN, we selected four hidden layers of the discriminative network $D$ to calculate the perceptual adversarial loss. We next proceed to analyze the property of these hidden layers.  Specifically, we trained four configurations of the PAN to perform the task of transforming the semantic labels to the cityscapes images. For each configuration, we set one hyper-parameter $\lambda_i$ as 1, and the others $\{\lambda_{1},\dots,\lambda_{i-1},\lambda_{i+1},\dots\}$ as 0, \ie, we used only one hidden layer to evaluate the perceptual adversarial loss in each configuration. As shown in Fig.~\ref{fig:analysis}, the lower layers (\eg, $\ell_{percep}^{D, 1}$, $\ell_{percep}^{D, 2}$) pay more attention to the patch-to-patch transformation and the color transformation, but the transformed images are blurry and lack of fine details. On the other hand, higher layers (\eg, $\ell_{percep}^{D, 1}$, $\ell_{percep}^{D, 2}$) capture more high-frequency information, but lose the color information. 
Therefore, by integrating different properties of these hidden layers, the proposed PAN can be expected to achieve better performance, and the final results of this task are shown in Fig.~\ref{cityscapes2} and Table~\ref{table:pix2pix}.

In our work, the balance between the perceptual adversarial loss and GAN loss is controlled by the hyper-parameters $\theta$. In the task of transforming labels to facades, we vary the value of $\theta$ to test its influence on the proposed PAN. Qualitative samples are reported in Fig~\ref{facades}. As shown in Fig.~\ref{facades}, only using the perceptual adversarial loss (\ie, $\theta=0$) has already had the capability of synthesizing visually reasonable images from the input labels. Given the advantage of the GAN loss to promote more realistic images, the transformation performance gets better with increasing $\theta$. However, with the continuous increasing of $\theta$, the role of perceptual adversarial loss will be weakened and the model performance drops, \eg, visual artifacts are observed in certain images.

\subsection{Comparing with existing works}

In this subsection, we compared the performance of the proposed PAN with those of existing algorithms for image-to-image transformation tasks.

\subsubsection{Context-encoder}

\begin{table}[!t]
\renewcommand{\arraystretch}{1.3}
\caption{De-raining}
\label{table:deraining}
\centering
\begin{tabular}{|c|c|c|c|c|}
\hline
  & PSNR(dB) & SSIM & UQI & VIF \\
\hline
\hline
 L2 & 22.77 & \ \ 0.7959 \ \  & \ \ 0.6261 \ \ &  \ \ 0.3570 \ \  \\
\hline
 cGAN & 21.87 & \ \ 0.7306 \ \  & \ \ 0.5810 \ \ &  \ \ 0.3173 \ \  \\
\hline
 L2+cGAN & 22.19 & 0.8083 & 0.6278 & 0.3640 \\
\hline
 ID-CGAN & 22.91 & 0.8198 & 0.6473 & 0.3885\\
\hline
 PAN & {\bf 23.35} & {\bf 0.8303} & {\bf 0.6644} & {\bf 0.4050}\\
\hline
 PA Loss+cGAN & 23.22 & 0.8078 & 0.6375 & 0.3904\\
\hline
\end{tabular}
\end{table}

\begin{table}[!t]
\renewcommand{\arraystretch}{1.3}
\caption{In-painting}
\label{table:inpainting}
\centering
\begin{tabular}{|c|c|c|c|c|}
\hline
  & PSNR(dB) & SSIM  & UQI & VIF\\
\hline
\hline
 Context-Encoder &  21.74  & \ \  0.8242 \ \  & \ \ 0.7828 \ \  & \ \ 0.5818 \ \  \\
\hline
 PAN & {\bf 21.85} & {\bf 0.8307} & {\bf 0.7956} & {\bf 0.6104} \\
\hline
\end{tabular}
\end{table}

Context-Encoder (CE)~\cite{pathak2016context} trained CNNs for the single image inpainting task. Given corrupted images as input, image inpainting can be formulated as an image-to-image transformation task. Pixel-wise loss (least squares loss) and the generative adversarial loss were combined in the Context-Encoder to explore the relationship between the input surroundings and its central missed region.

To compare with the Context-Encoder, we applied PAN to inpaint images whose central regions were missed. As illustrated in Section~\ref{sec:setup}, 100k images were randomly selected from the ILSVRC'12 dataset to train both Context-Encoder and PAN, and 50k images from the ILSVRC'12 validation set were used for test purpose. Moreover, since the image inpainting models are asked to generate the missing region of the input image instead of the whole image, we employ the image transformation network architecture from~\cite{pathak2016context}.

\begin{figure*}[!t]
\centering
\includegraphics[width=7.1in]{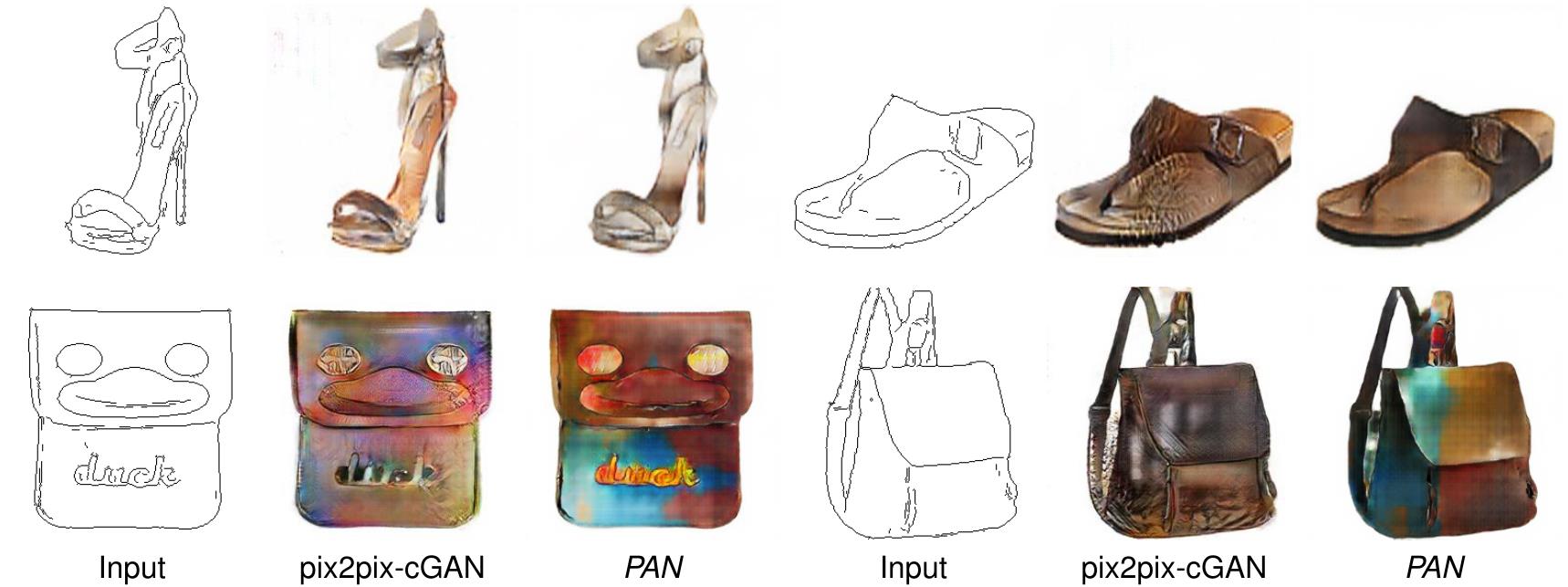}
\caption{Comparison of transforming the object edges to corresponding images using the pix2pix-cGAN with the proposed PAN. Given the edges (leftmost), the generated images of shoes and handbags are listed on the rightside.}
\label{edges} 
\end{figure*}

\begin{figure*}[!t]
\centering
\includegraphics[width=7.1in]{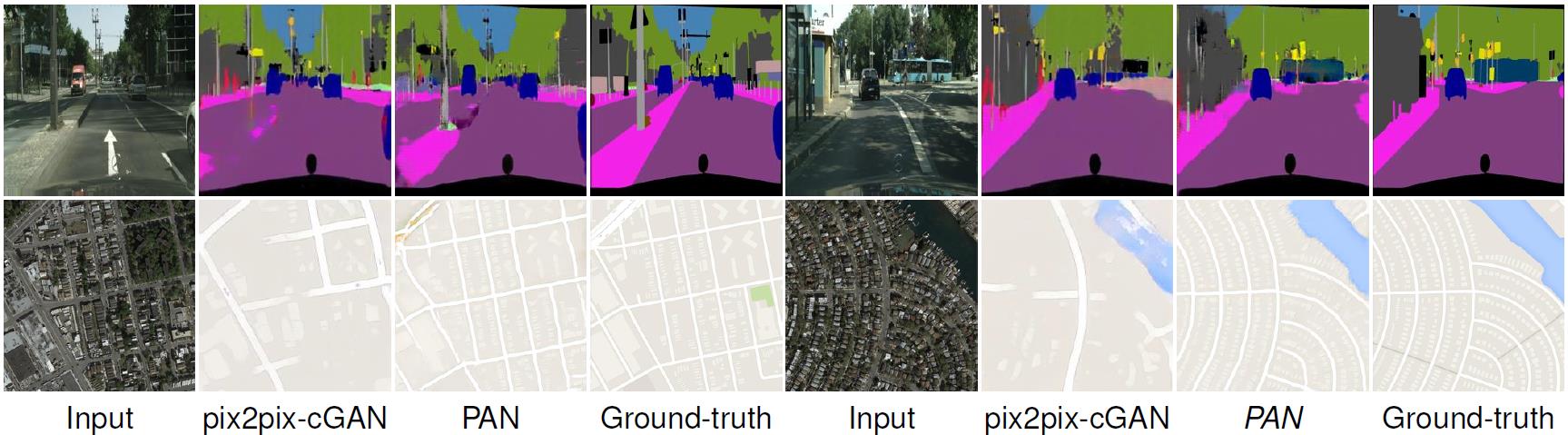}
\caption{Comparison of some other tasks using the pix2pix-cGAN with the proposed PAN. In the first row, semantic labels are generated based on the real-world cityscapes images. And, the second row reports the generated maps given the aerial photos as input.}
\label{maps} 
\end{figure*}

In Fig.~\ref{fig_sim}, we reported some example results in the test set. For each input image, the missing part is mixed by the foreground objects and backgrounds. From the inpainted results, we find the proposed PAN performed better on understanding the surroundings and estimating the missing part with semantic contents. However, the context-encoder tended to use the nearest region (usually the background) to inpaint the missing part. PAN can synthesize more details in the missing parts. Last but not the least, in Table~\ref{table:inpainting}, we reported the quantitative results calculated over all 50k test images, which also demonstrated that the proposed PAN achieves better performance. 

\subsubsection{ID-CGAN}

Image de-raining task aims to remove rain streaks in a given rainy image. Considering the unpredictable weather conditions, the single image de-raining (de-snowing) is a challenge image-to-image transformation task. Most recently, the Image De-raining Conditional Generative Adversarial Networks (ID-CGAN) was proposed to tackle the image de-raining problem. Through combining the pixel-wise (least squares loss), conditional generative adversarial, and perceptual losses (VGG-16), ID-CGAN achieved the state-of-the-art performance on single image de-raining. 

We attempted to solve image de-raining by the proposed PAN using the same setting with that of ID-CGAN. Since there is a lack of large-scale datasets consisting of paired rainy and de-rained images, we resort to synthesize the training set~\cite{zhang2017image} of 700 images. Zhang~\etal~\cite{zhang2017image} provided 100 synthetic images and 50 real-world rainy images for the test. Since the ground-truth is available for synthetic test images, we calculated and reported the quantitative results in Table~\ref{table:deraining}. Moreover, we test both ID-CGAN and PAN on real-world rainy images, and the results were shown in Fig.~\ref{derain_idcgan}. For better visual comparison, we zoomed up the specific regions-of-interest below the test images.

From Fig.~\ref{derain_idcgan}, we found both ID-CGAN and PAN achieved great performance on single image de-raining. However, by observing the zoomed region, the PAN removed more rain-strikes with less color distortion. Additionally, as shown in Table~\ref{table:deraining}, for synthetic test images, the de-rained results of PAN are much more similar with the corresponding ground-truth than that of ID-CGAN. Dealing with the uncontrollable weather condition, why the proposed PAN can achieve better results? One possible reason is that ID-CGAN utilized the well-trained classifier to extract the high-level features of the output and ground-truth images, and penalize the discrepancy between them (\ie, the perceptual loss). The high-level features extracted by the well-trained classifier usually focus on the content information, and may hard to capture other image information, such as color information. Yet, the proposed PAN used the perceptual adversarial loss, which aims to continually and automatically measure the discrepancy between the output and ground-truth images. The different training strategy of PAN may help the model to learn a better mapping from the input to output images, and resulting in better performance. 

\begin{figure*}[!t]
\centering
\includegraphics[width=\textwidth]{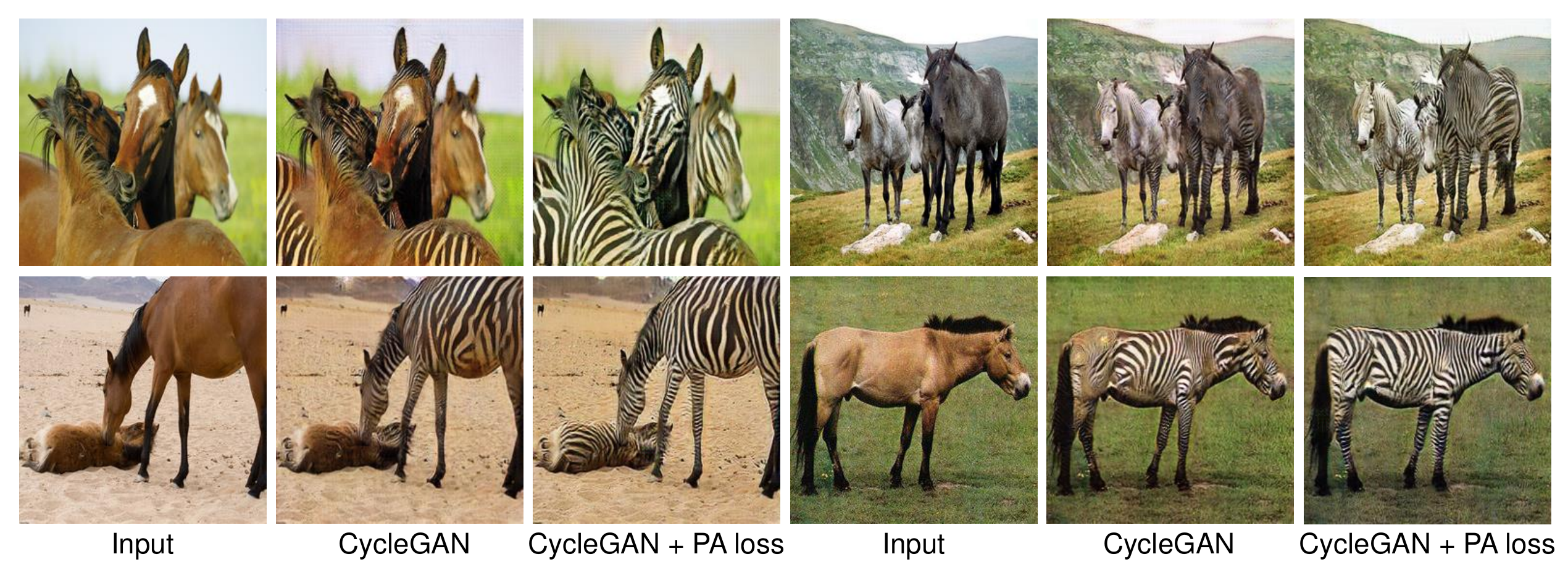} \vskip -0.15 in
\caption{Introducing the proposed perceptual adversarial loss to CycleGAN framework and attempting to perform unpaired image translation (horses$\leftrightarrow$zebras). Given horse images as input, trained models (both CycleGAN and `CycleGAN+Perceptual adversarial loss') aim to generate corresponding zebras images.}
\label{horses2zebras} 
\end{figure*}

\subsubsection{Pix2pix-cGAN}

\begin{table}[!t]
\renewcommand{\arraystretch}{1.3}
\caption{Comparison wiht pix2pix-CGAN}
\label{table:pix2pix}
\centering
\begin{tabular}{|c|c|c|c|c|}

\multicolumn{5}{c}{Senmatic labels $\to$ Cityscapes imags}  \\
\hline
 & PSNR(dB) & SSIM & UQI & VIF \\
\hline
\hline
  pix2pix-cGAN & 15.74  & \ \ 0.4275 \ \ & \ 0.07315 \ & \ 0.05208 \ \\
\hline
 PAN & {\bf 16.06} & {\bf 0.4820} & {\bf 0.1116} & {\bf 0.06581} \\

\hline
\multicolumn{5}{c}{Edges $\to$  Shoes} \\
\hline
 & PSNR(dB) & SSIM & UQI & VIF \\
\hline
\hline
 ID-cGAN & {\bf 20.07} & 0.7504 & 0.2724 & 0.2268 \\
\hline
 PAN & 19.51 & {\bf 0.7816} & {\bf 0.3442} & {\bf 0.2393}\\

\hline
\multicolumn{5}{c}{Edges $\to$  Handbags} \\
\hline
 & PSNR(dB) & SSIM & UQI & VIF \\
\hline
\hline
 ID-cGAN & {\bf 16.50} & 0.6307 & 0.3978 & 0.1723\\
\hline
 PAN & 15.90 & {\bf 0.6570} & {\bf 0.4042} & {\bf 0.1841} \\
\hline
\multicolumn{5}{c}{Cityscapes images $\to$  Semantic labels} \\
\hline
 & PSNR(dB) & SSIM & UQI & VIF \\
\hline
\hline
 ID-cGAN & 19.46 & 0.7270 & 0.1555 & 0.1180\\
\hline
 PAN & {\bf 20.67} & {\bf 0.7725} & {\bf 0.1732} & {\bf 0.1638}\\
\hline
\multicolumn{5}{c}{Aerial photos $\to$  Maps} \\
\hline
 & PSNR(dB) & SSIM & UQI & VIF \\
\hline
\hline
 ID-cGAN & 26.10 & 0.6465 & 0.09125 & 0.02913\\
\hline
 PAN & {\bf 28.32} & {\bf 0.7520} & {\bf 0.3372} & {\bf 0.1617} \\
\hline
\end{tabular}
\end{table}

Isola~\etal~\cite{isola2016image} utilized cGANs as a general-purpose solution to image-to-image translation (transformation) tasks. In their work, the pixel-wise loss (least absolute loss) and Patch-cGANs loss are employed to solve a serial of image-to-image transformation tasks, such as translating the object edges to its photos, semantic labels to scene images, gray images to color images, \etc. The proposed PAN can also solve the image-to-image transformation tasks performed by pix2pix-cGAN. Here, we implemented some of them and compared with pix2pix-cGAN.  

Firstly, we attempted to translate the semantic labels to cityscapes images. Unlike the image segmentation problems, this inverse translation is an ill-posed problem and image transformation network has to learn prior knowledge from the training data. As shown in Fig.~\ref{cityscapes2}, given semantic labels as input images, we listed the transformed cityscapes images of pix2pix-cGAN, PAN and the corresponding ground-truth on the rightside. From the comparison, we found the proposed PAN captured more details with less deformation, which led the synthetic images are looked more realistic. Moreover, the quantitative comparison in Table~\ref{table:pix2pix} also indicated that the PAN can achieve much better performance.

Generating real-world objects from corresponding edges is also one kind of image-to-image transformation task. Based on the dataset provided by~\cite{isola2016image}, we trained the PAN to translate edges to object photos, and compared its performance with that of pix2pix-cGAN. Given edges as input, Fig.~\ref{edges} presented shoes and handbags synthesized by pix2pix-cGAN and PAN. At the same time, the quantitative results over the test set were shown in the Table~\ref{table:pix2pix}. Observing the generated object photos, we think that both pix2pix-cGAN and PAN achieved promising performance, yet it's hard to tell which one is better. Quantitative results are also very close, PAN performed slightly inferior to pix2pix-cGAN on the PSNR measurement, yet superior on other quantitative measurements. 

In addition, we compared PAN with pix2pix-cGAN on tasks of generating semantic labels from cityscapes photos, and generating maps from the aerial photos. Some example images generated using PAN and pix2pix-cGAN and their corresponding quantitative results were shown in Fig.~\ref{maps} and Table~\ref{table:pix2pix}, respectively. To perform these two tasks, the image-to-image transformation models are asked to capture the semantic information from the input image, and synthesize the corresponding transformed images. Since pix2pix-cGAN employed the pixel-wise and generative adversarial losses to training their model, it may hard to capture perceptual information from the input image, which causes that their results are poor, especially on transforming the aerial photos to maps. However, we can observe that the proposed PAN can still achieve promising performance on these tasks. These experiments showed that the proposed PAN can also effectively extract the perceptual information from the input image.

Overall, in the cityscapes dataset (Fig.~\ref{cityscapes2}), semantic labels correspond to different kinds of objects, such as vehicle, road, tree, \etc. Meanwhile, objects in the same category usually share some common patterns and features. Similarly, in the aerial2map dataset (Fig.~\ref{maps}), both aerial images and maps have their own shared patterns, which is beneficial for learning the mapping relation between them. In contrast, in the edges2images task (Fig.~\ref{edges}), given a handbag sketch, it can correspond to hundreds of kinds of outputs with different colors, textures, \etc. Therefore, compared with Fig.~\ref{cityscapes2} (cityscapes) and Fig.~\ref{maps} (aerial2map), transformation relations in Fig.~\ref{edges} (edges2images) are more difficult and challenging, which leads to relatively small visual improvement.

\subsection{Extension to unpaired image translations}

As discussed in section~\ref{sec:background}, besides learning paired image-to-image transformations, some works~\cite{Zhu_2017_ICCV,Yi_2017_ICCV,pmlr-v70-kim17a} investigated cross-domain image translations and performed image translations in absence of paired examples. The proposed perceptual adversarial loss aims to continually explore and minimize the discrepancy between perceptual features of generated images and that of the target images. However, in unpaired image translation tasks, the target image corresponding to a generated image is unknown. Alternatively, given training samples $\{x_i\in\mathcal{X}\}^{N}_{i=1}$ and $\{y_i\in\mathcal{Y}\}^{N}_{i=1}$ in two domains, we calculate the discrepancy of mean features on two domains,
\begin{equation}
\begin{aligned} 
\ell_{percep}^{D, j} = ||\mathbb{E}_{\mathcal{Y}}d_j(y_i) - \mathbb{E}_{\mathcal{X}}d_j(G(x_i)) ||,
\end{aligned} 
\end{equation} 
where $G(\cdot)$ and $d_j (\cdot)$ represent the output of generators and the representation on the j\textsuperscript{th} hidden layer of discriminators in the CycleGAN framework, respectively. 

In this section, we perform the unpaired image translation task, horse$\leftrightarrow$zebra, and qualitatively report some generated results. For a fair comparison, we adopted default settings (and codes) from CycleGAN, and utilized the 3\textsuperscript{rd} and 4\textsuperscript{th} hidden layers of the discriminator to measure the perceptual adversarial loss. In addition, hyper-parameters $\lambda_3=\lambda_4=0.5$, $m=0.1$, and batch size of 4 were used. As shown in Fig. x, through considering the perceptual similarity in unpaired image translations, the model performance was more or less improved. Although this work mainly focuses on exploring the perceptual features between paired images (the generated image and its ground-truth), we demonstrate the possibility of improving the performance of unpaired image translations through measuring the perceptual similarity between different image domains.

\section{Conclusion}
\label{sec:conclusion}

In this paper, we proposed the perceptual adversarial networks (PAN) for image-to-image transformation tasks. As a generic framework of learning mapping relationship between paired images, the PAN combines the generative adversarial loss and the proposed perceptual adversarial loss as a novel training loss function. According to this loss function, a discriminative network $D$ is trained to continually and automatically explore the discrepancy between the transformed images and the corresponding ground-truth images. Simultaneously, an image transformation network $T$ is trained to narrow the discrepancy explored by the discriminative network $D$. Through the adversarial training process, these two networks are updated alternately. Finally, experimental results on several image-to-image transformation tasks demonstrated that the proposed PAN framework is effective and promising for practical image-to-image transformation applications.

\section*{Acknowledgment}
The authors would like to thank the handling associate editor Dr. Catarina Brites and all anonymous reviewers for their positive support and constructive comments for improving the quality of this paper.

\bibliographystyle{IEEEtran}

\bibliography{bare_jrnl}

\begin{thebibliography}{10}
\providecommand{\url}[1]{#1}
\csname url@samestyle\endcsname
\providecommand{\newblock}{\relax}
\providecommand{\bibinfo}[2]{#2}
\providecommand{\BIBentrySTDinterwordspacing}{\spaceskip=0pt\relax}
\providecommand{\BIBentryALTinterwordstretchfactor}{4}
\providecommand{\BIBentryALTinterwordspacing}{\spaceskip=\fontdimen2\font plus
\BIBentryALTinterwordstretchfactor\fontdimen3\font minus
  \fontdimen4\font\relax}
\providecommand{\BIBforeignlanguage}[2]{{%
\expandafter\ifx\csname l@#1\endcsname\relax
\typeout{** WARNING: IEEEtran.bst: No hyphenation pattern has been}%
\typeout{** loaded for the language `#1'. Using the pattern for}%
\typeout{** the default language instead.}%
\else
\language=\csname l@#1\endcsname
\fi
#2}}
\providecommand{\BIBdecl}{\relax}
\BIBdecl

\bibitem{elad2006image}
M.~Elad and M.~Aharon, ``Image denoising via sparse and redundant
  representations over learned dictionaries,'' \emph{IEEE Transactions on Image
  processing}, vol.~15, no.~12, pp. 3736--3745, 2006.

\bibitem{bertalmio2000image}
M.~Bertalmio, G.~Sapiro, V.~Caselles, and C.~Ballester, ``Image inpainting,''
  in \emph{Proceedings of the 27th annual conference on Computer graphics and
  interactive techniques}.\hskip 1em plus 0.5em minus 0.4em\relax ACM
  Press/Addison-Wesley Publishing Co., 2000, pp. 417--424.

\bibitem{nasrollahi2014super}
K.~Nasrollahi and T.~B. Moeslund, ``Super-resolution: a comprehensive survey,''
  \emph{Machine vision and applications}, vol.~25, no.~6, pp. 1423--1468, 2014.

\bibitem{luan2007natural}
Q.~Luan, F.~Wen, D.~Cohen-Or, L.~Liang, Y.-Q. Xu, and H.-Y. Shum, ``Natural
  image colorization,'' in \emph{Proceedings of the 18th Eurographics
  conference on Rendering Techniques}.\hskip 1em plus 0.5em minus 0.4em\relax
  Eurographics Association, 2007, pp. 309--320.

\bibitem{khan2014survey}
M.~W. Khan, ``A survey: Image segmentation techniques,'' \emph{International
  Journal of Future Computer and Communication}, vol.~3, no.~2, p.~89, 2014.

\bibitem{fu2017clearing}
X.~Fu, J.~Huang, X.~Ding, Y.~Liao, and J.~Paisley, ``Clearing the skies: A deep
  network architecture for single-image rain removal,'' \emph{IEEE Transactions
  on Image Processing}, vol.~26, no.~6, pp. 2944--2956, 2017.

\bibitem{pathak2016context}
D.~Pathak, P.~Krahenbuhl, J.~Donahue, T.~Darrell, and A.~A. Efros, ``Context
  encoders: Feature learning by inpainting,'' in \emph{The IEEE Conference on
  Computer Vision and Pattern Recognition (CVPR)}, 2016.

\bibitem{dong2016image}
C.~Dong, C.~C. Loy, K.~He, and X.~Tang, ``Image super-resolution using deep
  convolutional networks,'' \emph{IEEE transactions on pattern analysis and
  machine intelligence}, vol.~38, no.~2, pp. 295--307, 2016.

\bibitem{zhang2016colorful}
R.~Zhang, P.~Isola, and A.~A. Efros, ``Colorful image colorization,'' in
  \emph{European Conference on Computer Vision (ECCV)}, 2016.

\bibitem{cheng2015deep}
Z.~Cheng, Q.~Yang, and B.~Sheng, ``Deep colorization,'' in \emph{The IEEE
  International Conference on Computer Vision (ICCV)}, 2015.

\bibitem{shelhamer2016fully}
E.~Shelhamer, J.~Long, and T.~Darrell, ``Fully convolutional networks for
  semantic segmentation,'' \emph{IEEE transactions on pattern analysis and
  machine intelligence}, 2016.

\bibitem{du2018quantum}
Y.~Du, T.~Liu, Y.~Li, R.~Duan, and D.~Tao, ``Quantum divide-and-conquer
  anchoring for separable non-negative matrix factorization,'' \emph{arXiv
  preprint arXiv:1802.07828}, 2018.

\bibitem{goodfellow2014generative}
I.~Goodfellow, J.~Pouget-Abadie, M.~Mirza, B.~Xu, D.~Warde-Farley, S.~Ozair,
  A.~Courville, and Y.~Bengio, ``Generative adversarial nets,'' in
  \emph{Advances in neural information processing systems (NIPS)}, 2014, pp.
  2672--2680.

\bibitem{miyato2018cgans}
T.~Miyato and M.~Koyama, ``cgans with projection discriminator,'' in
  \emph{Proceedings of the International Conference on Learning Representations
  (ICLR)}, 2018.

\bibitem{isola2016image}
``Image-to-image translation with conditional adversarial networks.''

\bibitem{Zhu_2017_ICCV}
J.-Y. Zhu, T.~Park, P.~Isola, and A.~A. Efros, ``Unpaired image-to-image
  translation using cycle-consistent adversarial networks,'' in \emph{The IEEE
  International Conference on Computer Vision (ICCV)}, 2017.

\bibitem{Yi_2017_ICCV}
Z.~Yi, H.~Zhang, P.~Tan, and M.~Gong, ``Dualgan: Unsupervised dual learning for
  image-to-image translation,'' in \emph{The IEEE International Conference on
  Computer Vision (ICCV)}, Oct 2017.

\bibitem{pmlr-v70-kim17a}
T.~Kim, M.~Cha, H.~Kim, J.~K. Lee, and J.~Kim, ``Learning to discover
  cross-domain relations with generative adversarial networks,'' in
  \emph{Proceedings of the 34th International Conference on Machine Learning
  (ICML)}, vol.~70, 06--11 Aug 2017, pp. 1857--1865.

\bibitem{chen2018attention}
X.~Chen, C.~Xu, X.~Yang, and D.~Tao, ``Attention-gan for object transfiguration
  in wild images,'' \emph{arXiv preprint arXiv:1803.06798}, 2018.

\bibitem{johnson2016perceptual}
J.~Johnson, A.~Alahi, and L.~Fei-Fei, ``Perceptual losses for real-time style
  transfer and super-resolution,'' in \emph{European Conference on Computer
  Vision (ECCV)}.\hskip 1em plus 0.5em minus 0.4em\relax Springer, 2016, pp.
  694--711.

\bibitem{dosovitskiy2016generating}
A.~Dosovitskiy and T.~Brox, ``Generating images with perceptual similarity
  metrics based on deep networks,'' in \emph{Advances in Neural Information
  Processing Systems (NIPS)}, 2016, pp. 658--666.

\bibitem{bruna2015super}
J.~Bruna, P.~Sprechmann, and Y.~LeCun, ``Super-resolution with deep
  convolutional sufficient statistics,'' in \emph{Proceedings of the
  International Conference on Learning Representations (ICLR)}, 2016.

\bibitem{simonyan2014very}
K.~Simonyan and A.~Zisserman, ``Very deep convolutional networks for
  large-scale image recognition,'' \emph{arXiv preprint arXiv:1409.1556}, 2014.

\bibitem{zhang2017image}
H.~Zhang, V.~Sindagi, and V.~M. Patel, ``Image de-raining using a conditional
  generative adversarial network,'' \emph{arXiv preprint arXiv:1701.05957},
  2017.

\bibitem{ledig2016photo}
C.~Ledig, L.~Theis, F.~Husz{\'a}r, J.~Caballero, A.~Cunningham, A.~Acosta,
  A.~Aitken, A.~Tejani, J.~Totz, Z.~Wang \emph{et~al.}, ``Photo-realistic
  single image super-resolution using a generative adversarial network,'' in
  \emph{The IEEE Conference on Computer Vision and Pattern Recognition (CVPR)},
  2017.

\bibitem{zhang2018unreasonable}
R.~Zhang, P.~Isola, A.~A. Efros, E.~Shechtman, and O.~Wang, ``The unreasonable
  effectiveness of deep features as a perceptual metric,'' in \emph{The IEEE
  Conference on Computer Vision and Pattern Recognition (CVPR)}, 2018.

\bibitem{rumelhart1988learning}
D.~E. Rumelhart, G.~E. Hinton, and R.~J. Williams, ``Learning representations
  by back-propagating errors,'' \emph{Cognitive modeling}, vol.~5, no.~3, p.~1,
  1988.

\bibitem{eigen2013restoring}
D.~Eigen, D.~Krishnan, and R.~Fergus, ``Restoring an image taken through a
  window covered with dirt or rain,'' in \emph{The IEEE International
  Conference on Computer Vision (ICCV)}, 2013.

\bibitem{ruzic2015context}
T.~Ruzic and A.~Pizurica, ``Context-aware patch-based image inpainting using
  markov random field modeling,'' \emph{IEEE Transactions on Image Processing},
  vol.~24, no.~1, pp. 444--456, 2015.

\bibitem{qin2014novel}
C.~Qin, C.-C. Chang, and Y.-P. Chiu, ``A novel joint data-hiding and
  compression scheme based on smvq and image inpainting,'' \emph{IEEE
  transactions on image processing}, vol.~23, no.~3, pp. 969--978, 2014.

\bibitem{farabet2013learning}
C.~Farabet, C.~Couprie, L.~Najman, and Y.~LeCun, ``Learning hierarchical
  features for scene labeling,'' \emph{IEEE transactions on pattern analysis
  and machine intelligence}, vol.~35, no.~8, pp. 1915--1929, 2013.

\bibitem{noh2015learning}
H.~Noh, S.~Hong, and B.~Han, ``Learning deconvolution network for semantic
  segmentation,'' in \emph{The IEEE International Conference on Computer Vision
  (ICCV)}, 2015.

\bibitem{eigen2015predicting}
D.~Eigen and R.~Fergus, ``Predicting depth, surface normals and semantic labels
  with a common multi-scale convolutional architecture,'' in \emph{The IEEE
  International Conference on Computer Vision (ICCV)}, 2015.

\bibitem{yang2015weakly}
J.~Yang, S.~E. Reed, M.-H. Yang, and H.~Lee, ``Weakly-supervised disentangling
  with recurrent transformations for 3d view synthesis,'' in \emph{Advances in
  Neural Information Processing Systems (NIPS)}, 2015, pp. 1099--1107.

\bibitem{wang2017tag}
C.~Wang, C.~Wang, C.~Xu, and D.~Tao, ``Tag disentangled generative adversarial
  network for object image re-rendering,'' in \emph{Proceedings of the
  Twenty-Sixth International Joint Conference on Artificial Intelligence
  (IJCAI)}, 2017, pp. 2901--2907.

\bibitem{eigen2014depth}
D.~Eigen, C.~Puhrsch, and R.~Fergus, ``Depth map prediction from a single image
  using a multi-scale deep network,'' in \emph{Advances in neural information
  processing systems (NIPS)}, 2014, pp. 2366--2374.

\bibitem{chen2016infogan}
X.~Chen, Y.~Duan, R.~Houthooft, J.~Schulman, I.~Sutskever, and P.~Abbeel,
  ``Infogan: Interpretable representation learning by information maximizing
  generative adversarial nets,'' in \emph{Advances in Neural Information
  Processing Systems (NIPS)}, 2016, pp. 2172--2180.

\bibitem{zhao2016energy}
J.~Zhao, M.~Mathieu, and Y.~LeCun, ``Energy-based generative adversarial
  network,'' in \emph{Proceedings of the International Conference on Learning
  Representations (ICLR)}, 2017.

\bibitem{arjovsky2017wasserstein}
M.~Arjovsky, S.~Chintala, and L.~Bottou, ``{W}asserstein generative adversarial
  networks,'' in \emph{Proceedings of the 34th International Conference on
  Machine Learning (ICML)}, 2017.

\bibitem{gulrajani2017improved}
I.~Gulrajani, F.~Ahmed, M.~Arjovsky, V.~Dumoulin, and A.~Courville, ``Improved
  training of wasserstein gans,'' in \emph{Advances in Neural Information
  Processing Systems (NIPS)}, 2017.

\bibitem{karras2017progressive}
T.~Karras, T.~Aila, S.~Laine, and J.~Lehtinen, ``Progressive growing of gans
  for improved quality, stability, and variation,'' in \emph{Proceedings of the
  International Conference on Learning Representations (ICLR)}, 2018.

\bibitem{wang2018evolutionary}
C.~Wang, C.~Xu, X.~Yao, and D.~Tao, ``Evolutionary generative adversarial
  networks,'' \emph{arXiv preprint arXiv:1803.00657}, 2018.

\bibitem{miyato2018spectral}
T.~Miyato, T.~Kataoka, M.~Koyama, and Y.~Yoshida, ``Spectral normalization for
  generative adversarial networks,'' in \emph{Proceedings of the International
  Conference on Learning Representations (ICLR)}, 2018.

\bibitem{Li_2017_CVPR}
J.~Li, X.~Liang, Y.~Wei, T.~Xu, J.~Feng, and S.~Yan, ``Perceptual generative
  adversarial networks for small object detection,'' in \emph{The IEEE
  Conference on Computer Vision and Pattern Recognition (CVPR)}, July 2017.

\bibitem{lotter2015unsupervised}
W.~Lotter, G.~Kreiman, and D.~Cox, ``Unsupervised learning of visual structure
  using predictive generative networks,'' in \emph{Proceedings of the
  International Conference on Learning Representations (ICLR) Workshop}, 2016.

\bibitem{zhu2016generative}
J.-Y. Zhu, P.~Kr{\"a}henb{\"u}hl, E.~Shechtman, and A.~A. Efros, ``Generative
  visual manipulation on the natural image manifold,'' in \emph{European
  Conference on Computer Vision (ECCV)}.\hskip 1em plus 0.5em minus 0.4em\relax
  Springer, 2016, pp. 597--613.

\bibitem{brock2016neural}
A.~Brock, T.~Lim, J.~Ritchie, and N.~Weston, ``Neural photo editing with
  introspective adversarial networks,'' in \emph{Proceedings of the
  International Conference on Learning Representations (ICLR)}, 2017.

\bibitem{wang2017high}
T.-C. Wang, M.-Y. Liu, J.-Y. Zhu, A.~Tao, J.~Kautz, and B.~Catanzaro,
  ``High-resolution image synthesis and semantic manipulation with conditional
  gans,'' in \emph{The IEEE Conference on Computer Vision and Pattern
  Recognition (CVPR)}, 2018.

\bibitem{chen2017photographic}
Q.~Chen and V.~Koltun, ``Photographic image synthesis with cascaded refinement
  networks,'' in \emph{The IEEE International Conference on Computer Vision
  (ICCV)}, 2017.

\bibitem{gatys2015neural}
L.~A. Gatys, A.~S. Ecker, and M.~Bethge, ``A neural algorithm of artistic
  style,'' in \emph{The IEEE Conference on Computer Vision and Pattern
  Recognition (CVPR)}, 2016.

\bibitem{park2017transformation}
E.~Park, J.~Yang, E.~Yumer, D.~Ceylan, and A.~C. Berg,
  ``Transformation-grounded image generation network for novel 3d view
  synthesis,'' in \emph{The IEEE Conference on Computer Vision and Pattern
  Recognition (CVPR)}, 2017.

\bibitem{radford2015unsupervised}
A.~Radford, L.~Metz, and S.~Chintala, ``Unsupervised representation learning
  with deep convolutional generative adversarial networks,'' in
  \emph{Proceedings of the International Conference on Learning Representations
  (ICLR)}, 2015.

\bibitem{zeiler2011adaptive}
M.~D. Zeiler, G.~W. Taylor, and R.~Fergus, ``Adaptive deconvolutional networks
  for mid and high level feature learning,'' in \emph{The IEEE International
  Conference on Computer Vision (ICCV)}, 2011.

\bibitem{dumoulin2016guide}
V.~Dumoulin and F.~Visin, ``A guide to convolution arithmetic for deep
  learning,'' \emph{arXiv preprint arXiv:1603.07285}, 2016.

\bibitem{cordts2016cityscapes}
M.~Cordts, M.~Omran, S.~Ramos, T.~Rehfeld, M.~Enzweiler, R.~Benenson,
  U.~Franke, S.~Roth, and B.~Schiele, ``The cityscapes dataset for semantic
  urban scene understanding,'' in \emph{The IEEE Conference on Computer Vision
  and Pattern Recognition (CVPR)}, 2016.

\bibitem{yu2014fine}
A.~Yu and K.~Grauman, ``Fine-grained visual comparisons with local learning,''
  in \emph{The IEEE Conference on Computer Vision and Pattern Recognition
  (CVPR)}, 2014.

\bibitem{xie2015holistically}
S.~Xie and Z.~Tu, ``Holistically-nested edge detection,'' in \emph{The IEEE
  International Conference on Computer Vision (ICCV)}, 2015.

\bibitem{bergstra2010theano}
J.~Bergstra, O.~Breuleux, F.~Bastien, P.~Lamblin, R.~Pascanu, G.~Desjardins,
  J.~Turian, D.~Warde-Farley, and Y.~Bengio, ``Theano: A cpu and gpu math
  compiler in python,'' in \emph{Proc. 9th Python in Science Conf}, 2010, pp.
  1--7.

\bibitem{kingma2014adam}
D.~Kingma and J.~Ba, ``Adam: A method for stochastic optimization,'' in
  \emph{Proceedings of the International Conference on Learning Representations
  (ICLR)}, 2014.

\bibitem{wang2004image}
Z.~Wang, A.~C. Bovik, H.~R. Sheikh, and E.~P. Simoncelli, ``Image quality
  assessment: from error visibility to structural similarity,'' \emph{IEEE
  transactions on image processing}, vol.~13, no.~4, pp. 600--612, 2004.

\bibitem{995823}
Z.~Wang and A.~C. Bovik, ``A universal image quality index,'' \emph{IEEE Signal
  Processing Letters}, vol.~9, no.~3, pp. 81--84, March 2002.

\bibitem{sheikh2006image}
H.~R. Sheikh and A.~C. Bovik, ``Image information and visual quality,''
  \emph{IEEE Transactions on image processing}, vol.~15, no.~2, pp. 430--444,
  2006.

\bibitem{dong2014learning}
C.~Dong, C.~C. Loy, K.~He, and X.~Tang, ``Learning a deep convolutional network
  for image super-resolution,'' in \emph{European Conference on Computer Vision
  (ECCV)}, 2014.

\bibitem{kingma2013auto}
D.~P. Kingma and M.~Welling, ``Auto-encoding variational bayes,'' in
  \emph{Proceedings of the International Conference on Learning Representations
  (ICLR)}, 2014.

\end{thebibliography}

\end{document}